\documentclass{custom}

\usepackage{microtype}
\usepackage{graphicx}
\usepackage{booktabs} % for professional tables

\usepackage{hyperref}

\usepackage{amsmath, amssymb, amsthm}
\usepackage{authblk}
\usepackage{natbib}
\usepackage{tabularx}
\usepackage{tikz}
\usepackage{makecell}% http://ctan.org/pkg/makecell
\usepackage{multirow}
\usepackage[font=footnotesize,caption=false]{subfig}
\usepackage{algpseudocode}
\usepackage[english]{babel}
\usepackage{csquotes}
\usepackage{wrapfig}
\usepackage{enumitem}
\usepackage{float}

\renewcommand{\paragraph}[1]{{\vspace*{6pt}\noindent\bf{#1 }}}

\newcommand{\AND}{\wedge}

\newcommand{\calG}{\mathcal{G}}

\newcommand{\calE}{\mathcal{E}}
\newcommand{\calD}{\mathcal{D}}

\newcommand{\calM}{\mathcal{M}}

\newcommand{\calP}{\mathcal{P}}
\newcommand{\calL}{\mathcal{L}}
\newcommand{\calN}{\mathcal{N}}
\newcommand{\prob}{\mathbb{P}}

\newcommand{\sign}{\text{sgn}}
\newcommand{\signw}{\text{sgn}(\widehat{w})}

\newcommand{\vecw}{\mathbf{w}}
\newcommand{\vecx}{\mathbf{x}}

\newcommand{\vecg}{\mathbf{g}}
\newcommand{\vecu}{\mathbf{u}}

\newcommand{\vectheta}{\boldsymbol{\theta}}

\newcommand{\grad}{\nabla}
\newcommand{\grads}{\calG}

\definecolor{lightgreen}{RGB}{175,255,175}
\definecolor{lightred}{RGB}{255,150,150}
\definecolor{darkgrey}{RGB}{170,170,170}
\definecolor{lightorange}{RGB}{255,207,158}
\definecolor{orange}{RGB}{255,127,80}
\definecolor{darkgreen}{RGB}{50,127,0}
\definecolor{Blue}{RGB}{0,0,255}

\newcommand{\authcnote}[3]{
	\ifnum\authnotes=1
		\textcolor{#3}{ [{\textbf{#1 says:} #2]}}
	\fi
}

\newcommand{\ignore}[1]{\if{0} #1 \fi}
\newcommand{\nignore}[1]{\if{true} #1 \fi}

\makeatletter
  \def\SOUL@hlpreamble{%
  \setul{}{2.5ex}%         !!!change this value!!! default is 2.5ex
  \let\SOUL@stcolor\SOUL@hlcolor
  \SOUL@stpreamble
}
\makeatother

\newcommand{\reals}{\mathbb{R}}
\newcommand{\argmin}{\text{argmin}}
\newcommand{\argmax}{\text{argmax}}

\theoremstyle{definition}
\newtheorem{definition}{Definition}[section]

\theoremstyle{remark}

\usepackage{url}

\title{Hard-label Manifolds: Unexpected Advantages of Query Efficiency for Finding On-manifold Adversarial Examples}

\author[1]{Washington Garcia}
\author[2]{Pin-Yu Chen}
\author[3]{Somesh Jha}
\author[4]{Scott Clouse}
\author[1]{Kevin R. B. Butler}

\affil[1]{\normalsize Florida Institute for Cyber Security (FICS) Research, University of
 Florida \authorcr
\textit{\small \{w.garcia,butler\}@ufl.edu}\vspace{1mm}}
\affil[2]{\normalsize IBM Research \authorcr
\textit{\small pin-yu.chen@ibm.com}\vspace{1mm}}
\affil[3]{\normalsize University of Wisconsin-Madison and XaiPient \authorcr
\textit{\small jha@cs.wisc.edu}\vspace{1mm}}
\affil[4]{\normalsize Air Force Research Laboratory \authorcr
\textit{\small hsclouse@ieee.org}\vspace{1mm}}

\begin{document}

\maketitle

\begin{abstract}
% deep networks robust to noise is an open problem
% recent hard label attacks on image classification models have shown comparable performance to their first-order alternatives
% it is well known that in the zeroth-order setting, the adversary must efficiently search for the nearest decision boundary
% the SotA generally rely on the concept of superpixels to efficiently search through the data manifold. 
% it was recently shown that regular adversarial examples tend to leave the data manifold, and on-manifold examples are simply generalization errors
% in this paper we argue that query efficiency in the zeroth-order setting is related to the adversary's traversal through the data manifold. in particular the adversary must balance query efficiency with the example's distance to the manifold. We show that against robustly trained models, an efficient zeroth-order attack tends towards the manifold rather than away from it. 

Designing deep networks robust to adversarial examples remains an open problem. Likewise, recent zeroth order hard-label attacks on image classification models have shown comparable performance to their first-order, gradient-level alternatives. It was recently shown in the gradient-level setting that regular adversarial examples leave the data manifold, while their on-manifold counterparts are in fact generalization errors. In this paper, we argue that query efficiency in the zeroth-order setting is connected to an adversary's traversal through the data manifold. To explain this behavior, we propose an information-theoretic argument based on a \textit{noisy manifold distance oracle}, which leaks manifold information through the adversary's gradient estimate. Through numerical experiments of manifold-gradient mutual information, we show this behavior acts as a function of the effective problem dimensionality and number of training points. On real-world datasets and multiple zeroth-order attacks using dimension-reduction, we observe the same universal behavior to produce samples closer to the data manifold. This results in up to two-fold decrease in the manifold distance measure, regardless of the model robustness. Our results suggest that taking the manifold-gradient mutual information into account can thus inform better robust model design in the future, and avoid leakage of the sensitive data manifold.

%  In particular, query-efficient hard-label attacks have the unexpected advantage of finding adversarial examples close to the data manifold. 

% It is well known that in this setting, the adversary must search for the nearest decision boundary in a query-efficient manner. State-of-the-art (SotA) attacks rely on the concept of pixel grouping, or super-pixels, to perform efficient boundary search. 

% Further, when a normal zeroth-order attack is made query-efficient through the use of pixel grouping, it can make up to a two-fold increase in query efficiency, and in some cases, reduce a sample's distance to the manifold by an order of magnitude.
% \kbnote{It would help if we have something quantitative to say here}
\end{abstract}

% \PY{(This sentence does not seem totally right. Should we say "query-efficient hard-label attacks have unexpected advantage of finding adversarial examples closer to the data manifold"?)}

\section{Introduction}
\label{sec:intro}

% The introduction

% previous work~\citep{Stutz_2019_CVPR}
% profitability of query reduction is known (in score-level).
% offmanifold vs. onmanfiold examples - "data geometric" view of human alignment problem~\citep{engstrom_learning_2019}

Adversarial examples against deep learning models were originally investigated as blind spots in classification~\citep{szegedy_intriguing_2013,Goodfellow2014}. Formal methods for discovering these blind spots emerged, which we denote as gradient-level attacks, and became the first techniques to reach widespread attention within the deep learning community~\citep{Papernot2015,moosavi-dezfooli_deepfool:_2015,carlini_towards_2016,carlini_adversarial_2017}. In order to compute the necessary gradient information, such techniques required access to the model parameters and a sizeable query budget. These shortcomings were addressed by the creation of score-level attacks, which only require the confidence values output by the deep learning models~\citep{Fredrikson2015,Tramer2016,CPY17zoo,ilyas_black-box_2018}. However, these attacks still rely on models to divulge information that would  be impractical to receive in real-world systems. By contrast, hard-label attacks make no assumptions about receiving side information, making it the weakest yet most realistic threat model. These methods, which originated from a random-walk on the decision boundary~\citep{brendel_decision-based_2017}, have been carefully refined to offer convergence guarantees~\citep{cheng2018query}, query efficiency~\citep{chen_hopskipjumpattack:_2019,cheng_sign-opt_2020}, and capability in the physical world~\cite{feng_query-efficient_2020}.

\begin{wrapfigure}{R}{0.4\textwidth}
    \centering
    \includegraphics[width=0.35\textwidth,trim={0 0 0 0},clip]{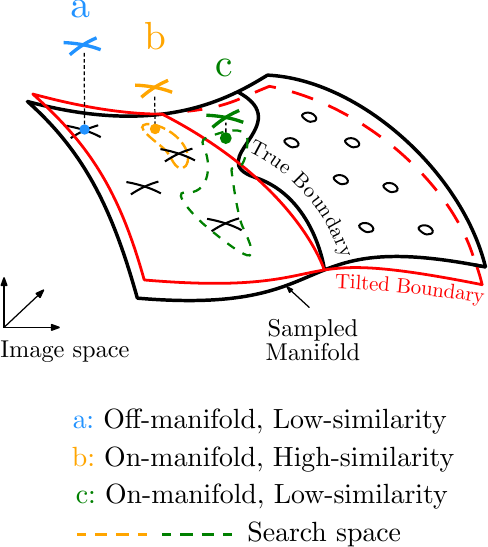}
    \caption{Our interpretation of zeroth-order attack behavior in the context of boundary tilting~\citep{tanay_boundary_2016}: a) zeroth-order attack targeting low-level features, leaving the manifold, b) an efficient zeroth-order attack targeting mostly high-level features, floating along the manifold, and c) manifold-based zeroth-order attack next to the manifold, but sacrificing similarity. For example, in our evaluation a, b, and c would correspond to Sign-OPT, BiLN+Sign-OPT, and AE+Sign-OPT, respectively.}
    \label{fig:visual_show}
\end{wrapfigure}

Despite the steady improvements of hard-label attacks, open questions persist about their behavior, and AML attacks at large. Adversarial samples were originally assumed to lie in rare pockets of the input space~\citep{Goodfellow2014}, but this assumptions was later challenged by the boundary tilting assumption~\citep{tanay_boundary_2016,gilmer_relationship_2018}, which adopts a ``data-geometric'' view of the input space living on a lower-dimensional manifold. This is supported by~\citet{Stutz_2019_CVPR}, who suggest that regular adversarial examples leave the data manifold, while on-manifold adversarial examples are generalization errors. From a data-geometric perspective, a sample's distance to the manifold primarily describes the amount of semantic features preserved during the attack process. This makes it advantageous to produce on-manifold adversarial examples, since the adversary can exploit the inherent generalization error of the model, while producing samples that are semantically similar. However, the true data manifold is either difficult or impossible to describe, and relying solely on approximations of the manifold can lead to the creation of crude adversarial examples~\citep{Stutz_2019_CVPR}.

In this paper, we adopt the boundary-tilting assumption and demonstrate an unexpected benefit of query-efficient zeroth order attacks, i.e., attacks primarily enabled by the use of down-scaling techniques. These attacks are more likely to discover on-manifold examples, which we argue is the result of manifold-gradient mutual information. While seemingly counter-intuitive, since down-scaling techniques reduce the search dimension (artificially limiting the search space for adversarial examples), our results suggest that the manifold-gradient mutual information can actually {\em increase} as a function of the effective dimensionality, and number of training points. This behavior leads to examples that are on-manifold generalization errors. With this knowledge, it is possible to rethink the design of hard-label attacks, to target high-level features as in attack (b) in Figure~\ref{fig:visual_show}, rather than (a) or (c). 

% \#1: unexpected benefit of query efficient search: find on manifold examples without approximating the manifold \PY{(Need to comment why it is unexpected. E.g., using super-pixel reduces the search dimension, so intuitively it seems to limit the search space and may find worse (further-away) adversarial examples). However, our results suggest that the side effect of imposing super-pixels is eliminating the search space of off-manifold adversarial examples, so it is unexpected.}

Our specific contributions are as follows:

\begin{itemize}[leftmargin=*]
    \item \textbf{Introduction of manifold distance oracle.} To create on-manifold examples, the adversary must leverage manifold information during the attack phase. We thus propose an information-theoretical formulation of the noisy manifold distance (NMD) oracle, which can explain how zeroth-order attacks craft on-manifold examples. We experimentally demonstrate that {\em manifold-gradient mutual information can increase as a function of the effective problem dimensionality and number of training points.} This finding relates to known behavior in the gradient-level setting, where manifold information can be leaked from robust models~\citep{engstrom_learning_2019}; unlike that work, however, our formulation describes leakage from both natural {\em and} robust models.
    % Studying this problem may assist in understanding the fundamental limits and utility of hard-label attacks.
    \item \textbf{Reveal new insights of manifold feedback during query-efficient zeroth-order search.} We describe an approach for extending dimension-reduction techniques in the score-level setting~\citep{tu_autozoom_2019} to hard-label attacks. We propose the use of FID score~\citep{heusel_gans_2018} as an $L_p$-agnostic means for estimating the adversary's manifold information gain. This methodology allows us to empirically demonstrate the connection between dimension reduction and manifold feedback from the model, beyond the known convergence rates tied to dimensionality~\citep{nesterov_random_2017}. 
    % (related to \#1) reveal a connection between query efficient attacks and on-manifold robustness in zeroth-order setting. propose use of FID score as measure of distance to the manifold for candidate samples. when searching on approximated manifold (with AE), it surpasses the normal variant for robust models, while BiLN variants can make an attack stay closer to manifold against natural models
    \item \textbf{Attack-agnostic method for super-pixel grouping.} We show that bilinear down-scaling of the input space act as a form of super-pixel grouping, yielding up to 28\% and 76\% query efficiency gain for previously-proposed HSJA~\citep{chen_hopskipjumpattack:_2019} and Sign-OPT attacks~\citep{cheng_sign-opt_2020}, respectively against robust models.More importantly, we show that super-pixel grouping produces samples close to the manifold and exploits the inherent generalization error of the model.

    % super-pixel method of query efficiency can be replicated by any attack with a simple bilinear transformation function. Show that this can also encourage sparsity in the generated adversarial noise. 
    % \item combination of bilinear transform with HSJA and sign-OPT surpasses their original variants on Imagenet, with BiLN HSJA outperforming or matching RayS in both L2 and Linf
    % \item twofold increase in query efficiency for SignOPT with BiLN
\end{itemize}

% An alternative approach uses zeroth-order optimization (ZOO), a technique that enables gradient approximation of a function given only query access to the function~\citep{ghadimi_stochastic_2013,nesterov_random_2017,liu_primer_2020}. Later treatment focused on improving the query efficiency of ZOO-based methods, leveraging dimensionality reduction of the gradient search space.

% focusing on first order attacks does not capture the entire picture

% searching over true high dimensional manifold yields better attacks (vertices of Linf norm ball), at the expense of query efficiency

% in high dimension there is low probability of moving in the exact direction of the decision boundary

% in lower dimension there is higher probability of moving along the decision boundary

% \kbnote{Itemize the contributions}

\section{Related Work}
\label{sec:relwork}

Since the original discovery of adversarial samples against deep models~\citep{szegedy_intriguing_2013,Goodfellow2014}, the prevailing question was why such examples existed. The original assumption was that adversarial examples lived in low-probability pockets of the input space, and were never encountered during parameter optimization~\citep{szegedy_intriguing_2013}. This effect was believed to be amplified by the linearity of weight activations in the presence of small perturbations~\citep{Goodfellow2014}. These assumptions were later challenged by the manifold assumption, which in summary 1) asserts that the train and test sets of a model only occupy a sub-manifold of the true data, while the decision boundary lies close to samples on and beyond the sub-manifold~\citep{tanay_boundary_2016}, and 2) supports the ``data geometric`` view, where high-dimensional geometry of the true data manifold enables a low-probability error set to exist~\citep{gilmer_relationship_2018}. Likewise the manifold assumption describes adversarial samples as leaving the manifold, which has inspired  defenses based on projecting such samples back to the data manifold~\citep{jalal_robust_2019,defensegan}. However, these approaches were later defeated by adaptive attacks~\citep{carlini_evaluating_2019,carlini_adversarial_2017,tramer_adaptive_2020}. We investigate the scenario where an adversary uses zeroth-order information to estimate the desired gradient direction~\citep{cheng_sign-opt_2020,chen_hopskipjumpattack:_2019}. Thus the adversary uses only the top-1 label feedback from their model query to synthesize samples. The desire for better query efficiency motivated the use of dimension reduction in hard-label attacks. However, to date it is not completely understood how this relates to traversal through the data manifold. We leverage previous results of the gradient-level setting~\citep{Stutz_2019_CVPR,engstrom_learning_2019} to formulate an explanation of manifold leakage during hard-label adversarial attacks.

\section{Noisy Manifold Distance Oracle}
\label{sec:approach}
% \textbf{Current argument (from first paragraph of Section~\ref{sec:discuss}):}

% From an information-theoretical perspective, the zeroth-order adversary observes the \textit{noisy gradient}, which is leaked as side information by each model decision.  Under this explanation, the decision feedback by the model is viewed as a noisy manifold distance (NMD) oracle. The improvement of AE+Sign-OPT on robust CIFAR-10 can be argued as a result of the NMD oracle improving as well. This can be shown using the data processing inequality~\citep{beaudry_intuitive_2012}: if $\textsc{MI}(\calM, \vecg)$ increases, then $\textsc{MI}(\calM, \ddot{\vecg})$ also increases, where $\textsc{MI}$ is mutual information, $\calM$ is the manifold, and $\ddot{\vecg}$ is the noisy gradient. In words, the quality of the noisy gradient depends on the quality of the model's loss landscape, which can more closely resemble the manifold under robust regularization. This means a higher quality loss landscape leads to a higher quality zeroth-order attack. Qualitative evidence of this effect can be observed in Section~\ref{app:visual_cifar10} of the Appendix. 

% \textbf{Proposed argument:}

\citet{santurkar_image_2019} demonstrate that the gradients of robust models have higher visual semantic alignment with the data compared to gradients of standard models. This suggests a reduction in uncertainty when sampling from the distribution of visual perturbations. If this reduced uncertainty can be attributed to leaked knowledge of the original data manifold, an adversary could exploit this fact and produce samples closer to the manifold.

This connects to the hard-label setting as follows. First, recall a standard result in data processing, which states that if three random variables form the Markov chain ${\displaystyle X\rightarrow Y\rightarrow Z}$, then their mutual information (MI) has the relation ${\displaystyle I(X;Y)\geqslant I(X;Z)}$~\citep{beaudry_intuitive_2012}. Now we have the data manifold $\calM=X$, the input gradient $\calG=Y$, and the noisy gradient from the black-box hard-label attack as $\ddot{\calG}=Z$. If $I(\calM,\calG)$ is larger for adversarially robust models, this may also suggest $I(\calM,\ddot{\calG})$ is larger. In the information theoretic sense, does this mean the gradients of adversarially robust models reveal more information about the training data than standard models? An immediate follow-up concern is whether other factors can influence the model to reveal this information, such as the problem dimensionality. \citet{schmidt_adversarially_2018} have shown that robust training requires additional data as a function of the data dimensionality. To test this hypothesis, we leverage the data model and results from ~\citet{schmidt_adversarially_2018} to derive an analytical solution for $I(\calG, \calM)$. This allows us to estimate the mutual information gain (or lack thereof) from adding extra training samples, as is the case in the adversarially robust setting, or reducing the effective problem dimensionality, as is common for hard-label attacks.

% If the hypothesis is true, why don't we see evasion attack or data stealing attack that are ``easier'' on adversarially robust models in empirical studies? Current empirical results all suggest that adversarial robust models are more difficult to attack, which is counterintuitive from data processing inequality. Does this contradiction suggest current black-box attacks are sub-optimal (not reaching the upper limit)? On the other hand, if the hypothesis is not true, i.e., better visual alignment does not suggest larger  $I(\calM,\ddot{\vecg})$, what can we say about the relation between the data manifold and the gradient of adversarially robust models? Finally, is there any information theoretical oracle query analysis that can be used for characterizing the efficiency of black-box attack, in terms of $I(\calM,\vecg)$, $I(\calM,\ddot{\vecg})$, and $I(\vecg,\ddot{\vecg})$? To test the hypothesis, we leverage the data model and results from ~\citet{schmidt_adversarially_2018} to derive an analytical solution for $I(\calM, \vecg)$. This allows to estimate the mutual information gain (or lack of) from adding extra training samples, as is the case in the adversarially robust setting. 

\paragraph{Data model and weights.} Recall the Gaussian mixture data model from \citet{schmidt_adversarially_2018}:

\begin{definition}

(Gaussian model). Let $\vectheta* \in \reals^d$ be the per-class mean vector and let $\sigma > 0$ be the variance parameter. Then the $(\vectheta*, \sigma)$-Gaussian model is defined by the following distribution over $(\vecx, y) \in \reals^d \times \{\pm 1\}$: First, draw a label $y \in \{\pm 1\}$ uniformly at random. Then sample the data point $\vecx \in \reals^d$ from $\calN(y \cdot \vectheta*, \sigma^2I)$.
\end{definition}

The difficulty of classification (i.e., linear separability) is controlled by the parameter $\sigma^2$ since we maintain $||\vectheta^*|| \simeq \sqrt{d}$. The data manifold is paramaterized for the Gaussian model by $(\vectheta^*, \sigma)$. 
% and can be described geometrically as the $d-$dimensional union of spheres $\calS := \calS_{+y} \union \calS_{-y}$ each with radius $\sigma$ and center $y \cdot \vectheta^*$.

% Let $\vectheta* \in \reals^d$ be the per-class mean vector and let $\sigma > 0$ be the variance parameter. Use the data model with input-label pairs $(\vecx \in \reals^d, y \in \reals)$ sampled from a distribution $\calD: \vecx \rightarrow y$. Specifically we rely on the Gaussian data model of \citet{schmidt_adversarially_2018} where

% \begin{equation*}
%     y \sim \{-1, +1\}, ~~ \vecx \stackrel{\text{i.i.d}}{\sim} \calN(\eta y, 1)
% \end{equation*}

% where $\calN$ is the Gaussian distribution with mean $\mathbf{\mu}=\eta y$ and variance $\sigma^2 = 1$. The separation between Gaussians is controlled by $\eta$ and likewise made large to ensure high standard accuracy. 

Next, recall a standard definition of classification error.

\begin{definition}
(Classification error). Let $\calP: \reals^d \times \{\pm1\} \rightarrow \reals$ be a distribution. Then the classification error $\beta$ of a classifier $f: \reals^d \rightarrow \{\pm1\}$ is defined as $\beta = \prob_{(\vecx, y)\sim\calP}[f(\vecx)\neq y]$.
\end{definition}

\textbf{Optimal classification weight (non-robust): } Fix the  $(\vectheta^*, \sigma)$-Gaussian model with $||\vectheta^*||_2 = \sqrt{d}$ and $\sigma \leq c\cdot d^{\frac{1}{4}}$, where c is a universal constant. \citet{schmidt_adversarially_2018} prove that for the linear classifier $f_\vecw: \reals^d \rightarrow \{\pm1\}$ defined as $f_\vecw(x) = \sign(\vecw \cdot \vecx)$, setting $\widehat{\vecw} = y \cdot \vecx$ (using a single tuple from the distribution) yields a linear classifier $f_{\widehat{\vecw}}$ with classification error of at most 1\%, with high probability. 

\citet{schmidt_adversarially_2018} later define a notion of robust classification error based on a bounded worst-case perturbation of input samples, which we defer to their paper for reference. 

% \begin{definition}
% (Robust classification error)~\citep{schmidt_adversarially_2018}. Let $\calP: \reals^d \times \{\pm1\} \rightarrow \reals$ be a distribution and let $\calB: \reals^d \rightarrow \power(\reals^d)$ be a perturbation set, a power set $\power$ of $\reals^d$. Then the $\calB$-robust classification error $\beta$ of a classifier $f: \reals^d \rightarrow \{\pm1\}$ is defined as $\beta = \prob_{(x,y) \sim \calP} [\exists x' \in \calB(x): f(x') \neq y]$.
% \end{definition}

\paragraph{Optimal classification weight (robust). } Now fix $\sigma \leq c_1d^{\frac{1}{4}}$ for the universal constant $c_1$, and samples $(x_1, y_1),\cdots,(x_n,y_n)$ drawn $i.i.d$ from the $(\vectheta^*, \sigma)$-Gaussian model with $||\vectheta^*|| = \sqrt{d}$. \citet{schmidt_adversarially_2018} prove that the weight setting  $\widehat{w} = \frac{1}{n}\sum^n_{i=1} y_ix_i$ yields an  $l^\epsilon_\infty$-robust classification error of at most 1\% for the linear classifier $f_{\widehat{w}}$ if

\begin{equation}\label{eq:robust_n}
    n \geq
    \begin{cases}
        1, & \text{for  }~ \epsilon \leq \frac{1}{4} d^{-\frac{1}{4}} \\
        c_2\epsilon^2\sqrt{d}, & \text{for  }~ \frac{1}{4}d^{-\frac{1}{4}} \leq \epsilon \leq \frac{1}{4}
    \end{cases}
     , % \right}
\end{equation}

for a universal constant $c_2$. We can leverage the weight settings as a function of $n$ and $d$ to give a closed form solution of mutual information. 

\subsection{Gradient-Manifold Mutual Information (MI)}

For simplicity fix $d=1$. Notice the classifier $\text{sgn}(\cdot)$ is discontinuous at $x=0$. Instead we consider the sub-gradient of the classifier at $x < 0$ and $x > 0$. In either case (non-robust or robust), the input sub-gradient for  $f_{\widehat{w}}(x')$ is defined as $\nabla_{x'}f_{\widehat{w}} = \signw$. Since the weights are Gaussian distributed with $\widehat{w}\sim\calN(\vectheta^*, \sigma)$, we can define the distribution of gradients as $\grads \sim \text{Bernoulli}\left(\prob_{\widehat{w}\sim\calN}\left[\widehat{w} \geq 0\right]\right)$. This allows defining the manifold-gradient point-wise joint probabilities case-wise, for the respective values under $g$ and $x$. We are concerned with the sub-gradient cases where $x > 0$ (denoted $x^+$) and $x < 0$ (denoted $x^-$) which correspond to the fixed values $g\in \{-1,1\}$.

Using the standard definition of the Gaussian distribution for both gradient and manifold, we can derive a closed form solution for the manifold-gradient mutual information (MI). The complete derivation of the joint and marginal probabilities can be found in Section~\ref{sec:supp_derivation} of the Appendix. We leverage Riemann approximation of the Gaussians over finite distance $\Delta_i = x_{i} - x_{i-1}$ for positive points in the Gaussian distribution $x_i^* \in [x_{i-1}, x_i]$, where $i\in \{1,\ldots,n\}$.

Fix $\sigma=cd^{\frac{1}{4}}$ for the non-robust case and $\sigma=c_1d^{\frac{1}{4}}$ for the robust case. We denote the sub-manifold sampled from the positive ($y=1$) and negative ($y=-1$) classes as $\calM^+$ and $\calM^-$, respectively. After simplifying due to symmetry, we have the closed form solution of manifold-gradient mutual information, based on the standard definition of mutual information from information theory~\citep{cover_elements_2006},

\begin{equation}\label{eq:MI:exact}
    \begin{aligned}
        I(\grads, \calM)_\epsilon &= 2\int_{\calM^+} p(1, x^+) \,\text{log}(\frac{p(1, x^+)}{p_\calG(1) p_\calM(x^+)}) \, dx^+ + 2\int_{\calM^+} p(-1, x^+) \,\text{log}(\frac{p(-1, x^+)}{p_\calG(-1)p_\calM(x^+)}) \, dx^+.
    \end{aligned}
\end{equation}

This leads to the Riemann approximation of Equation~\ref{eq:MI:exact} used in numerical experiments as

\begin{equation}\label{eq:MI}
\begin{aligned}
    I(\grads, \calM)_\epsilon &= \frac{2}{\sqrt{2\pi} \sigma^2} \sum^{||\calM^+||}_{i=1}   \text{exp}(-\frac{(x_i^*-\theta)^2}{2\sigma^2}) \cdot \beta^+_i \Delta_i + \frac{2}{\sqrt{2\pi} \sigma^2} \sum^{||\calM^+||}_{i=1}   \text{exp}(-\frac{(x_i^*+\theta)^2}{2\sigma^2}) \cdot \beta^-_i \Delta_i .
\end{aligned}
\end{equation}

% \begin{equation}\label{eq:MI}
% \begin{aligned}
%     I(\grads, \calM)_\epsilon &= 2\sum^{||\calM^+||}_{i=1} \frac{1}{\sqrt{2\pi} \sigma^2} \text{exp}(-\frac{(x_i^*-\theta)^2}{2\sigma^2}) \, \text{log}\left( \frac{\text{exp}(\frac{(x_i^*-\theta)^2}{2\sigma^2}}{\lambda_+ [\text{exp}(-\frac{(x_i^*-\theta)^2}{2\sigma^2}) + \text{exp}(-\frac{(x_i^*+\theta)^2}{2\sigma^2})] } \right) (x_i - x_{i-1}) \\
%     &+ 2\sum^{||\calM^+||}_{i=1} \frac{1}{\sqrt{2\pi} \sigma^2} \text{exp}(-\frac{(x_i^*+\theta)^2}{2\sigma^2}) \, \text{log}\left( \frac{\text{exp}(-\frac{(x_i^*+\theta)^2}{2\sigma^2}}{\lambda_- [\text{exp}(-\frac{(x_i^*-\theta)^2}{2\sigma^2}) + \text{exp}(-\frac{(x_i^*+\theta)^2}{2\sigma^2})] } \right) (x_i - x_{i-1}),
% \end{aligned}
% \end{equation}

where 

\begin{equation}
    \begin{aligned}
        \beta^+_i &= -\frac{(x_i^*-\theta)^2}{2\sigma^2} - \text{log}(\lambda_+) - \text{log}\left(\text{exp}(-\frac{(x_i^*-\theta)^2}{\sigma^2}) + \text{exp}(-\frac{(x_i^*+\theta)^2}{\sigma^2})\right), \\
        \beta^-_i &= -\frac{(x_i^*+\theta)^2}{2\sigma^2} - \text{log}(\lambda_-) - \text{log}\left(\text{exp}(-\frac{(x_i^*-\theta)^2}{\sigma^2}) + \text{exp}(-\frac{(x_i^*+\theta)^2}{\sigma^2})\right), \\
        \lambda_+ &= \frac{1}{\sqrt{2\pi}\sigma} \sum_{j=1}^{n}  \text{exp}\left( {-\frac{1}{2} \cdot \frac{(x_j^* - \theta)^2}{\sigma^2} } \right) \Delta_i, \\
        \lambda_- &= \frac{1}{\sqrt{2\pi}\sigma} \sum_{j=1}^{n}  \text{exp}\left( {-\frac{1}{2} \cdot \frac{(x_j^* + \theta)^2}{\sigma^2} } \right)\Delta_i,
    \end{aligned}
\end{equation}

for $\Delta_i = x_{i} - x_{i-1}$, positive $x_i^* \in [x_{i-1}, x_i]$ and $x_j^* \in [x_{j-1}, x_j]$. The values $\lambda_+$ and $\lambda_-$ resolve to the Riemann approximation of the marginal probabilities for $g=1$ and $g=-1$, respectively ($n$ given by $(\epsilon, d)$ setting). The $(\epsilon, d)$ setting determines the $n$ independent draws from the manifold for calculating the gradient marginal.

\subsection{Mutual information as a function of dimensionality}
\label{sec:approach:mi_d}

% The inequality in Equation~\ref{eq:robust_n} can be rewritten in terms of $d$ with bounded $n$ and fixed $\epsilon$ as

% \begin{equation}\label{eq:robust_d}
%     \begin{cases}
%         d \leq \left( \frac{1}{4\epsilon} \right)^4, &\text{when } n = 1 \, \text{(non-robust)}, \\
%         \left( \frac{1}{4\epsilon} \right)^4 \leq d \leq \left( \frac{n}{c_2\epsilon^2} \right)^2 ,  &\text{otherwise (robust).}
%     \end{cases}
% \end{equation}

Recall that $d$ controls $\sigma$ in Equation~\ref{eq:MI}, i.e., $\sigma=cd^{\frac{1}{4}}$ for non-robust case and $\sigma = c_1d^{\frac{1}{4}}$ for robust case. This represents a constant fraction of the probability mass in Equation~\ref{eq:MI}, which causes the mutual information to vary with $d$. To study this further, we run numerical experiments to estimate $I(\calG, \calM)$ as derived in Equation~\ref{eq:MI}, while varying the dimensionality $d$ against values of $c_2 \in \{1, 5, 10, 15\}$ and $\epsilon \in \{0.0, 0.125, 0.250\}$. Notice that the universal constant $c_2$ for robust weight setting controls the scaling of training points in Equation~\ref{eq:robust_n}, while $c$ and $c_1$ scale the allowed variance as a function of $d$. This means that $c$ and $c_1$ can be fixed (we use $0.5$) while $c_2$ and $d$ can be changed to study their effect. As in the data model presented earlier, we assume dimension co-independence, i.e., the covariance matrix $\mathbf{\Sigma}=\sigma I$. We also assume the manifold can be described with $n$ points, i.e., $||\calM|| = n$ defined by Equation~\ref{eq:robust_n}.

\begin{figure*}[]
    \centering
    \includegraphics[scale=1.0,trim={0 3.2em 0 0},clip]{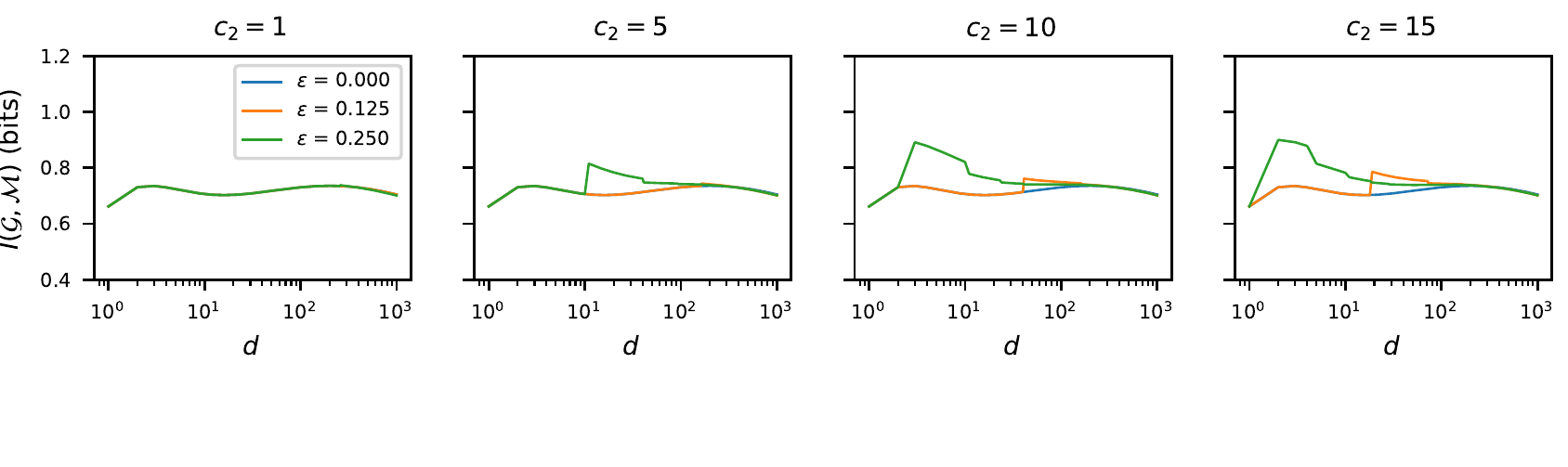}

    \caption{ a) Mutual information (MI) over dimension $d$ for several values of $c_2$ and $\epsilon$ in Equation~\ref{eq:MI:exact}, log-scale $d$-axis with $d \in [1, 10^3]$. The MI is higher at lower $d$ when $c_2$ (\# of training points) is increased w.r.t $d$, and when perturbation size $\epsilon$ is greater. }
    \label{fig:ItoD}
\end{figure*}

% Higher value of $c_2$, i.e. scaling up of training and manifold points w.r.t $d$, as well as higher $\epsilon$, allow for higher mutual information at a lower dimension.}

 The result of estimation is shown in Figure~\ref{fig:ItoD} with log-scale x-axis. When $c_2 = 1$, increased $\epsilon$ produces no change in mutual information. However, setting $c_2$ to higher values allows for higher mutual information from increased $\epsilon$. The effect is noticeable through to $c_2=15$. Given sufficient scaling factor $c_2$, mutual information increases reliably for lower dimension $d$. This effect is most noticeable when $\epsilon=0.25$ (green line). That is, given a sufficient surplus in training points, a robust model could act as an oracle leaking information through over-specification, i.e., include too many training points, or when the effective dimensionality of the learning problem is reduced. This can theoretically explain the high visual alignment observed empirically by \citet{engstrom_learning_2019} and \citet{santurkar_image_2019} on robust models. From an evasion adversary's perspective, the model cannot be modified to become over-specified, so this axis of mutual information gain is left for future work. Instead, we focus on the effective dimensionality, which the adversary can control through manifold-descriptive queries to the model, as it relates to higher values of $\epsilon$ in robust models. 
 
%  To summarize, we have shown that lowering the effective dimensionality allows for higher mutual information through the gradient, which can be leveraged by the adversary to produce samples closer to the manifold. 

% =========================================================
\section{Zeroth-order search through the manifold distance oracle}
\label{sec:approach2}

We showed from an information theoretic view that the true gradient during a black-box attack can act as a manifold distance oracle, and this oracle leaks more information based on the effective dimension of the learning problem. As a result, high MI between the true gradient and data manifold would suggest more information leakage between a noisy gradient estimate and the data manifold. We now investigate this phenomena in the context of real-world datasets. In the most common problem setting, the adversary is interested in attacking a $K$-way multi-class classification model $f: \reals^d \to \{1,\dots,K\}$. Given an original example $\vecx_0$, the goal is to generate adversarial example $\vecx$ such that $\vecx \text{ is close to } \vecx_0 \text{ and } f(\vecx) \neq f(\vecx_0),$ where closeness is often approximated by the $L_p$-norm of $\vecx - \vecx_0$. The value of this approximation is debated in the literature~\citep{heusel_gans_2018,tsipras_robustness_2018,engstrom_learning_2019}. We turn to alternative methods shown later for measuring closeness. First we step through the formulation for contemporary hard-label attacks, then show how dimension-reduced attacks can be formulated in the hard-label setting, which enables empirical analysis of our theoretical result.

% \label{eq:prob}

% \noindent\kbnote{Is the above sufficiently formal?}
% \noindent\wgnote{I took it from Cheng et al.'s paper (eq. 1) but we could change it } \PY{(We can say "where closeness is often approximated by the $\ell_p$ norm of $x-x_0$")}

\subsection{Gradient-level formulation}
\label{sec:approach:formulat}

For gradient-level attacks, the goal is satisfied by first assuming that $f(\vecx) = \argmax_i(Z(\vecx)_i)$, where $Z(\vecx) \in \reals^K$ is the final (logit) layer output, and $Z(\vecx)_i$ is the prediction score for the $i$-th class, the stated goal is satisfied by the optimization problem,

\begin{equation}\label{eq:wb}
    h(\vecx) := \underset{\vecx}{\operatorname{\argmin}} \left\{ ||\vecx - \vecx_0||_p + c\calL(Z(\vecx)) \right\}
\end{equation}

for the Euclidean $L_p$-norm $||\cdot||_p$, the loss function corresponding to the goal of the attack $\calL(\cdot)$, and a regularization parameter $c$. A popular choice of loss function is the \citet{carlini_towards_2016} loss function.

% \begin{equation}\label{eq:cwut}
%     \calL(Z(\vecx)) = \text{max} \left\{ \left[Z(\vecx) \right]_{y_0} - \max_{i\neq y_0} \left[ Z(\vecx)\right]_i, -\kappa \right\},
% \end{equation}

% \begin{equation}\label{eq:cwt}
%     \calL(Z(\vecx)) = \text{max} \left\{ \max_{i \neq t} \left[ Z(\vecx)\right]_i - Z(\vecx)_{t} , -\kappa \right\},
% \end{equation}

% where $y_0$ is the original label predicted by the classifier, and $\kappa$ is a parameter which controls ``confidence'' of the adversarial sample. 

% \subsection{Score-level formulation}

% In our investigation, we focus on zeroth-order methods which use a reconstruction of Equation~\ref{eq:wb}, assuming $h(\vecx)$ exists at every $\vecx$, with symmetric difference quotient method to approximate 

% \begin{equation}
%     \vecg_i = \frac{h(\vecx + h\vece_i) - h(\vecx - \mu\vece_i)}{2\mu} \approx \frac{\partial f(\vecx)}{\partial \vecx},
% \end{equation}

% for small $\mu \in \reals$ and the $i$-th elementary basis vector $\vece_i$. Likewise it is possible to adopt the attack objective $\calL$ in Equation~\ref{eq:cwt} but with $\kappa=0$.

\subsection{Score-level and hard-label attacks}
\label{sec:approach:scorelabel}

 In the gradient-level setting, we require the gradient $\grad f(\cdot)$. However, in the score-level setting we are forced to estimate $\frac{\partial f(\vecx)}{\partial \vecx}$ without access to $\grad f(\cdot)$, only evaluations of $Z(\cdot)$. \citet{tu_autozoom_2019} reformulate the previous problem to a version relying instead on the ranking of class predictions from $Z$. In practical scenarios, the estimate is found using random gradient-free method (RGF), a scaled random full gradient estimator of $\grad f(\vecx)$, over $q$ random directions $\{\vecu_i\}^q_{i=1}$. The score-level setting was extended to several renditions of the hard-label setting, which we clarify below. In each case the goal is to approximate the gradient by $\hat{\vecg}$.

% \begin{equation}\label{eq:zoom1}
%     h_{M}(\vecx) := \underset{\vecx}{\operatorname{\min}} \left\{ ||\vecx - \vecx_0||_p + c\calL\left(M[Z(\vecx)]\right) \right\},
% \end{equation}

% where $M(\cdot)$ is an entry-wise monotonic transformation which preserves the ranking of class predictions from $Z$. In the new form, it is 

% \begin{equation}\label{eq:zoom_est}
%     \vecg = \frac{1}{q} \sum_{j=1}^q b \cdot \frac{h_{M}(\vecx + \beta\vecu_j) - h_{M}(\vecx)}{\beta} \cdot \vecu_j,
% \end{equation}

% for smoothing parameter $\beta > 0$, unit-length random Gaussian vector $\vecu$, and tunable scaling parameter $b \in \reals$. 

\paragraph{OPT-Attack} For given example $\vecx_0$, true label $y_0$, and hard-label black-box function $f: \reals^d \to \{1, \dots, K\}$, \citet{cheng2018query} define the objective function $g: \reals^d \to \reals$ as a function of search direction $\vectheta$, where $g(\vectheta^*)$ is the minimum distance from $\vecx_0$ to the nearest adversarial example along the direction $\vectheta$. For the untargeted attack, $g(\vectheta)$ corresponds to the distance to the decision boundary along direction $\vectheta$, and allows for estimating the gradient as

\begin{equation}\label{eq:g_hat}
    \hat{\vecg} = \frac{1}{q} \sum^q_{i=0} \frac{g(\vectheta + \beta\vecu_i) - g(\vectheta)}{\beta} \cdot \vecu_i,
\end{equation}

where $\beta$ is a small smoothing parameter. Notably, $g(\vectheta)$ is continuous even if $f$ is a non-continuous step function. 

% However, $\grad g(\vectheta)$ cannot be evaluated exactly, only the values of $g$ using the hard-label queries to $f$. 

\paragraph{Sign-OPT} \citet{cheng_sign-opt_2020} later improved the query efficiency by only considering the sign of the gradient estimate, 

$$ \hat{\nabla} g(\vectheta) \approx \hat{\vecg} := \sum_{i=1}^q \sign\left(g(\vectheta + \beta\vecu_i) - g(\vectheta)\right) \vecu_i.  $$
% \begin{equation}
    
% \end{equation}

We focus on the Sign-OPT variant, since the findings are more relevant to the current state-of-the-art. 

\paragraph{HopSkipJumpAttack} Similar to Sign-OPT, HopSkipJumpAttack (HSJA)~\citep{chen_hopskipjumpattack:_2019} uses a zeroth-order sign oracle to improve Boundary Attack~\citep{brendel_decision-based_2017}. HSJA lacks the convergence analysis of OPT Attack/Sign-OPT and relies on one-point gradient estimate. Regardless, HSJA is competitive with Sign-OPT for state-of-the-art in the $L_2$ setting. 

Alternative hard-label attacks exist which do not rely on the explicit zeroth-order gradient estimate from the model, such as RayS by~\citet{chen_rays_2020}. We expect such attacks to behave differently, and thus provide results for RayS in Section~\ref{app:rays} of the Appendix.

% The attack proceeds in two stages:

% \begin{enumerate}
%     \item Exploration: use $\vecx = \vecx_0 + \calD(\vecz)$ and apply estimator from Equation~\ref{eq:zoom_est} with $q=1$ to an optimizer (ADAM) to solve Equation~\ref{eq:zoom1} until initial attack is found.
%     \item Exploitation: Continue to fine-tune adversarial perturbation $\calD(\vecz)$ for solving Equation~\ref{eq:zoom1} while setting $q\geq 1$ in Equation~\ref{eq:zoom_est}.
% \end{enumerate}

\subsection{Dimension-reduced zeroth-order search}
\label{sec:approach:dimreduce}

% The attacks described so far each represent an improvement in query efficiency under different $L_p$-norm scenarios. The difficulty in performing a holistic analysis of their behavior lies in each attack's unique design. 

In order to characterize hard-label attacks against the MI to dimension relationship shown in Section~\ref{sec:approach:mi_d}, we modify existing hard-label attacks to produce dimension-reduced variants. This scheme can allow dynamic scaling of the effective dimensionality up or down in a controlled manner. Our dimension-reduced search is feasible since the intrinsic dimensionality of data can be lower than the true dimension~\citep{amsaleg_vulnerability_2017}. In practice we implement the reduction through an encoding map $\calE: \reals^{d} \to \reals^{d'}$ for reduced dimension $d'$ and decoding map $\calD: \reals^{d'} \to \reals^{d}$. In general the adversarial sample is created by

\begin{equation}\label{eq:sample}
    \vecx = \vecx_0 + g\left(\calD(\vectheta')\right)\frac{\calD(\vectheta')}{||\calD(\vectheta')||},    
\end{equation}
 
where $\vectheta' \in \reals^{d'}$ and is optimized depending on the respective attack (e.g., Sign-OPT and HSJA), and as before, $g$ is a measure of distance to the decision boundary in direction $\calD(\vectheta')$. The mapping functions can be initialized with either an autoencoder (AE), or a pair of channel-wise bilinear transform functions (henceforth referred to as BiLN) which simply scales the input up or down depending on a fixed scaling factor. This represents two distinct methods of synthesizing adversarial samples, which either rely on an approximate description of the manifold (AE), or instead exploit the known spatial codependence of images (BiLN). 

% The inclusion of BiLN is important, since it allows to measure the scenario where the adversary has no explicit knowledge of the manifold, and only relies on the feedback from the model. 

% These choices were previously investigated by \citet{tu_autozoom_2019} as a way to improve query efficiency in score-level attacks, and ultimately performed similarly with respect to query efficiency. 

% \citet{tu_autozoom_2019} experiment in the score-level setting with a search over reduced-dimension directions $\vectheta' \in \reals^{d'}$. The motivation for this stems from the known convergence rate of zeroth order gradient descent, which is tied to a dimensionality $d$ of the vectorized input. The convergence rate is $O(\sqrt{d/T})$ for number of iterations $T$, on both convex~\citep{nesterov_random_2017,liu_zeroth-order_2018} and non-convex loss functions~\citep{ghadimi_stochastic_2013}. 

The adversary's AE is tuned to minimize reconstruction error of input images, so the output quality of the AE will depend on the adversary's ability to collect data. We assume the adversary only has access to the test set, which tends to be considerably less informative than the training set. This crude manifold approximation can manifest as an additional layer of distortion on top of adversarial noise. With BiLN, no additional training is required, so it synthesizes search directions independent of the adversary's manifold description (i.e., possible extracted knowledge about test samples). The complete implementation details of the AE variant can be found in Section~\ref{app:imp_details} of the Appendix. Next we describe how the mapping functions are used in our experiments.

% Under the AE scenario, the AE is tuned to minimize reconstruction error of input images. Due to this dependence on labeled data, the output quality of the AE is dependent on the adversary's ability to collect data, which is a realistic consideration. We model the scenario where the adversary only has access to the test set, which tends to be considerably less informative than the training set. This manifests as an extra distortion in addition to the adversarial noise. Thus the output of the AE-initialized decoder can be used in different ways, which we discuss briefly in Section~\ref{app:imp_details} of the Appendix. With BiLN, no additional training is required, which means it synthesizes search directions independent of the adversary's manifold description (i.e., possible extracted knowledge about test samples). This can manifest as a lower overall distortion, in the case where only a crude manifold description can be extracted from the test set. Next we describe the exact usage of the mapping functions for each attack scheme.

% Unfortunately, the trade-off between query efficiency and the manifold's fidelity is not well understood. 

% \revtwo{In the lower dimension case, the search takes place in a compressed lower fidelity version, which induces a penalty when projecting back to image space with $\calD$. Since the true manifold is not known, }

\paragraph{Sign-OPT \& HSJA.} In general, for attacks relying on the \citet{cheng2018query} formulation, the update in Equation~\ref{eq:g_hat} becomes

\begin{equation}\label{eq:ghat_update}
    \hat{\vecg} = \frac{1}{q} \sum^q_{i=0} \frac{g(\vectheta' + \beta\vecu'_i) - g(\vectheta')}{\beta} \cdot \vecu'_i
\end{equation}

for the reduced-dimension Gaussian vectors $\{\vecu'_i \in \reals^{d'}\}_{i=0}^{q}$ for integer $d' < d$ and direction $\vectheta' \in \reals^{d'}$. The reduced-dimension direction $\vectheta'$ is initialized randomly with $\vectheta'\sim\calN(0, 1)$ for the untargeted case, or for the targeted case as $\vectheta' = \calE(\vecx_t)$, where $\vecx_t$ is a test sample correctly classified as target class $t$ by the victim model. This scheme also applies to HSJA, since HSJA performs a single-point sign estimate. As in the normal variants, $\hat{\vecg}$ is used to update $\vectheta'$. 

% TODO: include SotA BiLN HSJA improvements
% \todo{signposting - reduced dim search improves efficiency in specific attacks}

% \kbnote{We may want some signposting through a few of these sections so that it's clear what our contribution is going to be}

\subsection{Estimating manifold-gradient mutual information}
\label{sec:approach:deviation}

We can leverage a sample's distance to the manifold as a signal of the effective gradient-manifold mutual information. Hereafter, we refer to this distance \textit{w.l.o.g} as the manifold distance. Unfortunately, the real data manifold is difficult to describe. This is an open problem in the study of Generative Adversarial Networks (GANs), since designers require that generator images are on-manifold to preserve semantic relationships between images. This has motivated the recently proposed Fréchet Inception Distance (FID) that acts as a surrogate measure of the manifold distance~\citep{heusel_gans_2018}. We can leverage FID by treating the adversarial samples as synthetically generated images, which are later compared to their unmodified counterparts on the true manifold. Since FID uses an Inception-V3 coding layer~\citep{szegedy2016rethinking} to encode images, this distance correlates with distortion of semantic high-level features. Thus sampling closer to the data manifold will result in a lower FID score. We do not target the Inception-V3 network in any of our experiments, so the FID metric will not rely on any internal aspects of the victim models.

% An advantage of manifold distance is it can communicate information about attack behavior better than $L_p$-norm measurements, which are common in existing zeroth-order attack literature~\citep{cheng_sign-opt_2020,chen_hopskipjumpattack:_2019,chen_rays_2020}. 

% Second, the manifold approximation technique by \citet{Stutz_2019_CVPR} is mainly suited for $L_2$-norm, whereas hard-label attacks exist for both $L_2$ and $L_{\infty}$-norm. 

% Stutz rely on L2 projection centered at test sample with nearest neighbor search to approximate manifold. To study this in multiple Lp norm settings, we use upper layers of Inception network, coupled with Frechet Inception Distance as a surrogate of the sample manifold distance. 

\section{Results}
\label{sec:results}

\subsection{Methodology} Our experimental analysis addresses the following three research questions about zeroth-order attacks:

\begin{enumerate}
    \item[Q1.] Under the analytical result of Section~\ref{sec:approach:mi_d}, does the adversary take advantage of increased manifold-gradient mutual information in practice?
    \item[Q2.] Similarly, is manifold-gradient mutual information affected by the model robustness?
    % \item[Q3.] Compared to the previous results in the score-level setting~\citep{tu_autozoom_2019}, do dimension-reduced hard-label attacks produce a similar amount of reduction in query usage?
    \item[Q3.] If the dimension is reduced as in Figure~\ref{fig:ItoD}, what is the trade-off between query efficiency and the resulting reduced search resolution?
\end{enumerate}

We study these questions by comparing two hard-label attacks with their compatible dimension-reduced variants, against both natural and robust models. The robust models contain an improved loss landscape as a result of increased sample complexity during the training or inference process. 

% Some variants are not shown due to incompatibility with the base attack. For example, AE+HSJA is not implemented as it relies on only a single-point estimate, thereby only allowing to attack on the manifold directly. This is not practical due to induced distortion discussed previously in Section~\ref{sec:approach2}. RayS can perform two-point search, but it assumes codependence of input features, which may not be the case for well-defined latent space of an autoencoder.\footnote{Experimentally, codependence hurt the AE+RayS variant more than was practically useful.} Thus for the AE variant, we rely on Sign-OPT as it can perform two-point estimate and does not rely on codependence of features. In the BiLN cases, the implementations follow the discussion in Section~\ref{sec:approach:dimreduce}.

% \PY{(To facilitate reading, I suggest we do "AE+HSJA" instead of "AE HSJA". Same applies to other combinations)}

\paragraph{Experimental Highlights.} Our experiments show that query-efficient attacks exhibit unexpected behaviors and benefits, with explanations summarized below:

\begin{enumerate}
    \item[A1.] Query-efficient gradient estimates reduce the effective dimensionality of the adversary's search. In experiments against CIFAR-10, this increases manifold-gradient mutual information, which allows lower FID-64 scores for BiLN+HSJA, BiLN+Sign-OPT, and AE+Sign-OPT. 
    % Likewise this is only observed on the gradient-estimate attacks, HSJA and Sign-OPT, while RayS showed little to no change. 
    \item[A2.] Robust models increase the sample complexity during training, which creates a cleaner loss landscape. Although we observe high FID-64 scores on the robust Madry model using normal attacks, dimension-reduction allows to match the low baseline scores of the natural model.
    % robust models have higher FID score regardless 
    % maybe existing attacks don't leverage the sample complexity
    \item[A3.] Dimension-reduced attacks are capable of state-of-the-art query-efficiency gains for HSJA and Sign-OPT against robust models, despite reducing the effective search resolution. 
\end{enumerate}

% generate 100 samples for each attack. FID score is calculated using the first 64-dimensional max pooling layer of Inception V3 (denoted as LID-64)~\footnote{\url{https://github.com/mseitzer/pytorch-fid}}. This allows to calculate FID without the full 2048 sample count of original FID, at the cost of losing some information. Should still be useful for comparing between attacks due to position of max pool in the network, although magnitudes will differ from those obtained by ~\cite{heusel_gans_2018}.

\paragraph{Setup.} All attacks run for 25k queries without early stopping. For brevity, we only show results for the untargeted case. FID score is calculated using the 64-dimensional max pooling layer of the Inception-V3 deep network for coding (denoted as FID-64), taken from an open-source implementation.\footnote{\url{https://github.com/mseitzer/pytorch-fid}} The choice of the 64-dimensional feature layer allows to calculate full-rank FID without the full 2,048 sample count of original FID, which is prohibitive based on the scale of our analysis. Since the coding layer differs slightly from the original FID-2048 implementation, the magnitudes will differ from those published by~\citet{heusel_gans_2018}.

% cifar10: case study - attack CIFAR10 using one convnet and three robust variants from the literature in the Linfty setting,

% mainly interested in zeroth order behavior when met with different levels of robustness to first-order distortion. 

Image data consists of the CIFAR-10~\citep{krizhevsky_learning_2009} classification dataset. Original samples are chosen from the test set using the technique from \citet{chen_hopskipjumpattack:_2019}: on CIFAR-10, ten random samples are taken from each of ten classes (i.e., 100 total samples). The natural CIFAR-10 network is the same implementation open-sourced by~\citet{cheng_sign-opt_2020}. In addition, we leverage the representative adversarial training technique proposed by~\citet{madry_towards_2017} (and their checkpoint) as the robust model. Additional results on the ImageNet dataset~\citep{ILSVRC15} can be found in Section~\ref{app:imagenet} of the Appendix. To avoid ambiguity, we label each BiLN variant with the spatial dimension after performing the bilinear transformation. 

% This selection allows to study attack behavior on both small and high-resolution image data. 

% Image data consists of the CIFAR-10~\citep{krizhevsky_learning_2009} and ImageNet~\citep{ILSVRC15} classification datasets. This selection allows to study attack behavior on both small and high-resolution image data. Original samples are chosen from the test set of each dataset using a similar technique as in \citet{chen_hopskipjumpattack:_2019}. On CIFAR-10, ten random samples are taken from each of ten classes. On ImageNet, ten random classes are chosen with ten random samples taken from each (i.e., 100 total samples on either dataset). We provide further implementation details in Section~\ref{app:imp_details} of the Appendix. In addition to natural images, we are interested in the attack behavior for a model regularized with some variant of first-order noise. For CIFAR-10, we choose the representative adversarial training technique proposed by~\citet{madry_towards_2017}. For completeness, we also include ablation results on more recent regularization techniques in Section~\ref{app:supp_results} of the Appendix. For ImageNet, we compare against the SotA at time of writing, randomized smoothing proposed by~\citet{cohen_certified_2019}. We use the pre-trained Resnet50 weights and implementation provided by Cohen et al., corresponding to smoothing parameter $\sigma=0.5$ and $\epsilon \simeq 1.0$. 

% \PY{(FID?)}
% ~\wgnote{0 step spike (targeted vs. untargeted)}

\subsection{Experimental details}
\label{sec:results:details}

We target the $L_p$-norm which the robust model was regularized under, i.e., $L_{\infty}$ versions of the attacks for CIFAR-10.

\begin{figure*}[t!]
    \centering
\subfloat[]{
    \includegraphics[scale=1.0,trim={0 0 0 0},clip]{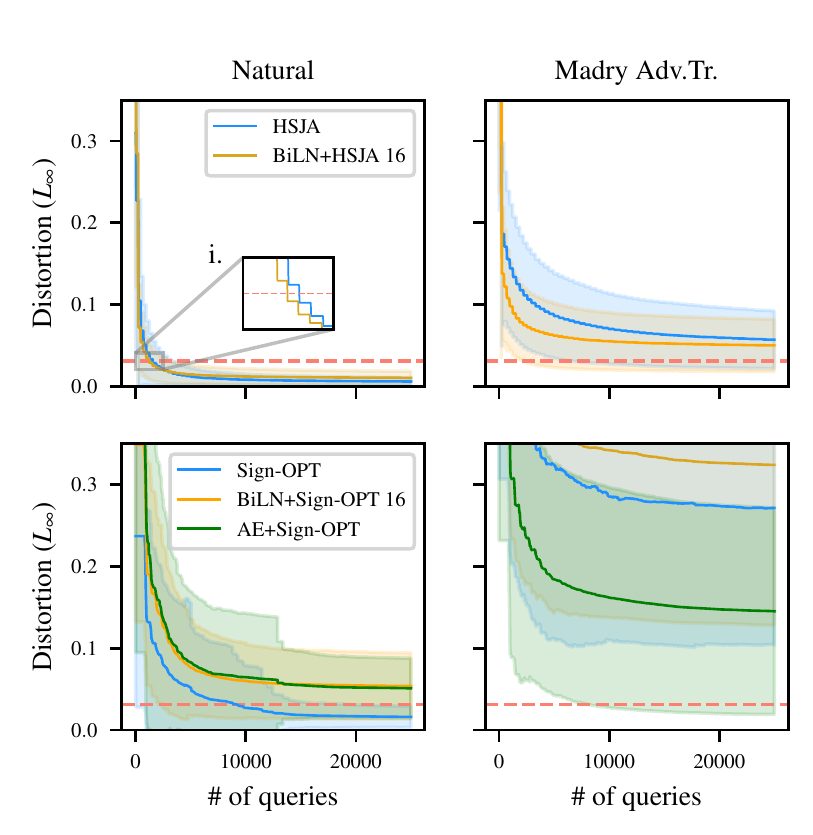}} 
% \hspace{0.2em}
\subfloat[]{
    \includegraphics[scale=1.0,trim={0 0 0 0},clip]{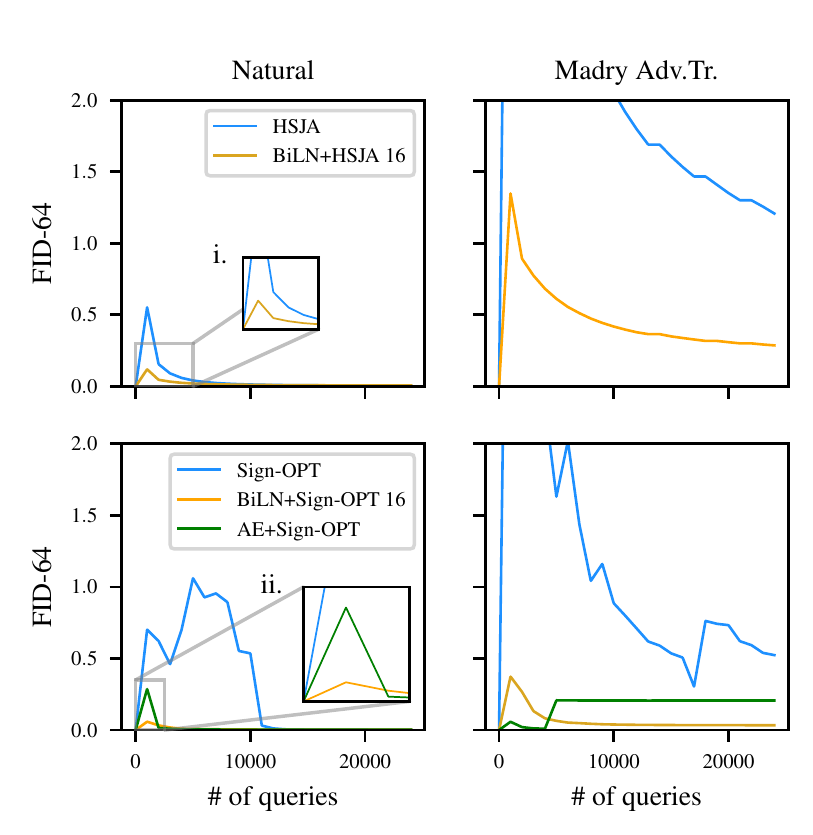}}\\
\vspace{0em}

% \vspace{-0.7em}
    \caption{Results across attacks for CIFAR-10 dataset, corresponding to a) distortion against query usage (dotted red line denotes the value of $\epsilon$, shaded areas mark standard deviation), and b) FID-64 trajectory against the same query usage. The estimate of manifold-gradient mutual information (FID-64) is universally lower on dimension-reduced attacks, such as BiLN and AE variants.}
    \label{fig:cifar10}
\end{figure*}

\paragraph{CIFAR-10 case study ($L_{\infty})$.} We measure the distortion against remaining query budget of the adversary in Figure~\ref{fig:cifar10}a. In general, the normal variants of each attack align with the published results. The main improvement is with BiLN+HSJA (orange line, top row) against the Madry adversarial training model, with average distortion at 4k queries decreasing from 0.09 to 0.07. This improvement is contrary to the minimal effect on the natural model (Inset~\ref{fig:cifar10}a.i blue line). AE+Sign-OPT (green line, bottom row) outperforms against regular Sign-OPT (blue line) and BiLN+Sign-OPT (orange line) on the robust model. However, the success with AE+Sign-OPT tends to be situational; in practice the low quality of the AE manifold description does not permit fine-grained adjustments to the perturbation. Overall, the Madry model could be weak against the zeroth-order distortions, since their formulation only considers first-order adversaries~\cite{madry_towards_2017}. To accompany the distortion results, we provide success rate plots for CIFAR-10 in Section~\ref{app:sr_plots} of the Appendix. 

Our next focus is Figure~\ref{fig:cifar10}b, which shows the FID score's trajectory over the search queries. Every trajectory will begin at a zero value, since there is an expected score of zero for identical images, and then peaks as the attack initialization is performed. Our main observation is the decrease in FID-64 score using the dimension-reduced variants. The magnitudes for AE+Sign-OPT (green line, Inset~\ref{fig:cifar10}b.ii) peaks at 0.28, then falls (and stays) near 0.004. BiLN+HSJA and BiLN+Sign-OPT (orange lines of Insets~\ref{fig:cifar10}b.i and \ref{fig:cifar10}b.ii) both exhibit lower FID scores than their normal variants (blue lines), as much as two orders of magnitude less in the case of BiLN+Sign-OPT. The reduced dimensionality leads to a lower FID-64 score, i.e., greater manifold information through the gradient estimate. On the Madry Adv. Tr. robust model (right column), we see a universal behavior where dimension-reduced variants produce a large reduction of FID-64 score. On Sign-OPT variants (bottom row), dimension reduction produces FID-64 score matching the baseline scores of the natural model. As a result, we can conclude that manifold information leakage occurs regardless of model sample complexity, and improves under certain dimension reduction schemes, such as BiLN+Sign-OPT and AE+Sign-OPT. We expand on this result in the next section. For completeness, we provide visual and tabular evidence of this behavior in Sections~\ref{app:visual_cifar10} and ~\ref{app:table_results_cifar10} of the Appendix, respectively.

\section{Discussion}
\label{sec:discuss}

% \PY{(How our results and the unexpected advantages inform future design of adversarial attack and defense? If one can have a more ideal "manifold description" (e.g., a better autoencoder), does it suggest better query efficient for ZO attack?  Will a better super-pixel grouping method yields better attack performance?) }

% \revtwo{\paragraph{Ablation against manifold projection.} }

%  In Section~\ref{sec:results} we observed that query-efficient attacks, i.e., those which leverage the concept of super-pixels to reduce search fidelity, are more likely to produce samples close to the manifold. This generates samples that are unseen by robust models during their first-order adversarial training. However, this takes place without any direct feedback about manifold distance. To approach this, we first consider that the model relies on a subsampled manifold of the image space. This manifold can be leaked by the loss landscape of the model, as shown by~\citet{engstrom_learning_2019}. 
 
\paragraph{Evidence of the noisy manifold distance oracle. } From an information-theoretical perspective, the zeroth-order adversary observes the \textit{noisy manifold distance} (NMD), which is leaked as side information by each gradient estimate. Against the natural model, dimension-reduced attacks led to a straightforward reduction in FID-64 score, as suggested by the numerical results of Section~\ref{sec:approach:mi_d}. Against the robust Madry model, normal attack variants had an inflated FID-64 score. However, BiLN+Sign-OPT and AE+Sign-OPT ultimately matched the baseline scores of the natural model. This suggests the NMD oracle improved in conjunction with the loss landscape of the victim model. This follows the data processing inequality (DPI)~\citep{beaudry_intuitive_2012}: if $\textsc{I}(\calM, \calG)$ increases, then $\textsc{I}(\calM, \ddot{\calG})$ also increases, where $\calM$ is the manifold, and $\ddot{\calM}$ is the noisy gradient. In words, the quality of the noisy gradient depends on the quality of the model's loss landscape, which can more closely resemble the manifold under robust regularization. This means a higher quality loss landscape leads to a higher quality zeroth-order attack. As we showed in Section~\ref{sec:approach}, this result is closely tied to the effective dimensionality, which can be arbitrarily lower than the true dimensionality~\cite{ma_characterizing_2018}.

% Qualitative evidence of this effect can be observed in Section~\ref{app:visual_cifar10} of the Appendix.  

\begin{table}[]
\centering
\begin{tabular}{lcc}
                                     & \multicolumn{2}{c}{Gradient Deviation}                                                                                                                          \\ \hline
\multicolumn{1}{|l|}{Attack Variant} & \multicolumn{1}{c|}{$||\vecg - \hat{\vecg}||_{\infty}$}                                                    & \multicolumn{1}{c|}{$||\vecg - \hat{\vecg}||_{2}$}                                                      \\ \hline
\multicolumn{1}{|l|}{Sign-OPT}       & \multicolumn{1}{c|}{\begin{tabular}[c]{@{}c@{}}4.154\\ $\pm$ 1.127\end{tabular}} & \multicolumn{1}{c|}{\begin{tabular}[c]{@{}c@{}}40.728\\ $\pm$ 9.166\end{tabular}}  \\ \hline
\multicolumn{1}{|l|}{\textbf{Sign-OPT+BiLN}}  & \multicolumn{1}{c|}{\begin{tabular}[c]{@{}c@{}}4.067\\ $\pm$ 1.252\end{tabular}} & \multicolumn{1}{c|}{\begin{tabular}[c]{@{}c@{}}40.643\\ $\pm$ 11.039\end{tabular}} \\ \hline
\multicolumn{1}{|l|}{Sign-OPT+AE}    & \multicolumn{1}{c|}{\begin{tabular}[c]{@{}c@{}}4.373\\ $\pm$ 1.118\end{tabular}} & \multicolumn{1}{c|}{\begin{tabular}[c]{@{}c@{}}49.587\\ $\pm$ 9.194\end{tabular}}  \\ \hline
\multicolumn{1}{|l|}{HSJA}           & \multicolumn{1}{c|}{\begin{tabular}[c]{@{}c@{}}4.884\\ $\pm$ 1.208\end{tabular}} & \multicolumn{1}{c|}{\begin{tabular}[c]{@{}c@{}}69.295\\ $\pm$ 6.332\end{tabular}}  \\ \hline
\multicolumn{1}{|l|}{HSJA+BiLN}      & \multicolumn{1}{c|}{\begin{tabular}[c]{@{}c@{}}4.584\\ $\pm$ 1.048\end{tabular}} & \multicolumn{1}{c|}{\begin{tabular}[c]{@{}c@{}}55.901\\ $\pm$ 6.963\end{tabular}}  \\ \hline
\end{tabular}
\caption{Comparison of the deviation between the true gradient and the first gradient estimate from various hard-label attack variants, in the direction of the successful adversarial prediction, averaged over 50 samples on CIFAR-10. }\label{tbl:gradients_compare}
\end{table}

\paragraph{Effect on gradient deviation.} When measuring FID-64 score, the score for BiLN+HSJA on the robust model failed to match the natural score, despite outperforming in distortion. A notable difference between HSJA and Sign-OPT (apart from analytic guarantees) is the method for performing estimates, e.g., one-point with HSJA and two-point for Sign-OPT. \citet{liu_primer_2020} showed that the one-point method can be noisier, which according to DPI will restrict the mutual information. We hypothesize that the one-point approach also leaves the manifold sooner, due to updating the reference sample on-the-fly. To quantify this, we calculate the $L_\infty$-norm and $L_2$-norm between the first gradient estimate for each attack variant and the true input gradient. The true input gradient is calculated from the original sample with respect to the cross-entropy loss between the model output and classification label of the adversarial sample. This comparison is shown in Table~\ref{tbl:gradients_compare} for the natural CIFAR-10 model, averaged over 50 samples from each attack. Sign-OPT variants have a universally lower gradient deviation than the HSJA variants. Notably, the Sign-OPT+BiLN variant (bolded) obtains the lowest gradient deviation, whereas HSJA has the highest deviation. These observations support our empirical argument that HSJA leads to a higher FID score, due to a higher variance in the gradient estimate as shown before by~\citet{liu_primer_2020}. The larger gradient deviation for HSJA implies noisier gradient estimation, and hence by the data processing inequality (DPI) leads to a lower mutual information (MI) between the data manifold and noisy gradients.

\paragraph{``Topology'' of hard-label settings.} We can view zeroth-order attacks as following a topological hierarchy that is a function of the effective data dimension. Our interpretation is illustrated in Figure~\ref{fig:visual_show}. Each technique offers a unique traversal distance both along the manifold, and away from it. Efficient attacks represented by (b) can combine elements of staying near manifold, and traversing it. This is representative of BiLN variants, which make basic assumptions about spatial correlation to balance search fidelity with manifold distance. In contrast, traversing close to an approximate manifold description with (c) introduces distortion as a result of the crude manifold description. Following the boundary tilting assumption, the nearest boundary on the manifold could also be far away. Thus we can consider an attack which learns a manifold description (e.g., an AE variant), but instead leverages the description to select the most relevant super-pixel grouping in the image. To this end, the FID score offers a reliable measure of manifold distance, which can inform the topological behavior, and the quality of future hard-label attacks. This ultimately enables a better evaluation of model robustness.  

% Alignment with \citet{Stutz_2019_CVPR}: only approach the classifier's decision boundary - 
% true decision boundary is not crossed for the zeroth-order scenario

% two-point estimate helps ensure the adversarial sample does not cross true decision boundary (label invariant defintion). of course there is no guarantee with class-agnostic autoencoders

\section{Conclusion}
\label{sec:conc}

Despite the recent progress in zeroth-order attack methods, open questions remain about their precise behavior. We develop an information-theoretic analysis that sheds light on their ability to produce on-manifold adversarial examples as a function of effective dimensionality. Through experiments on real-world datasets, we show up to two-fold decrease in the manifold distance by leveraging dimension-reduced attack variants. With knowledge of the manifold-gradient relationship, it is possible to further refine hard-label attacks, and inform a better evaluation of model robustness.

\section*{Acknowledgements}

This work was supported by the Air Force Office of Scientific Research (AFOSR) Grant FA9550-19-1-0169, and the National Science Foundation (NSF) Grants CNS-1815883 and  CNS-1562485. This work was partially supported by AFOSR Grant FA9550-18-1-0166, and NSF Grants CCF-FMitF-1836978, SaTC-Frontiers-1804648, CCF-1652140, and ARO grant number W911NF-17-1-0405.

\bibliography{bib_ml-attack,bib_ml-defense,bib_ml-interpret,bib_ml-privacy,bib_ml-surveys,bib_ml-theory,bib_ml-verify,bib_ml,bib_urls,bib_systems-sec,bib_infotheory}

\bibliographystyle{custom}

\appendix
\newpage
\section{Appendix}

\section{Derivation of Manifold-Gradient Mutual Information (MI)}
\label{sec:supp_derivation}

We define the manifold-gradient point-wise joint probability in a case-wise manner, for the respective values under $g$ and $x$. We are concerned with the sub-gradient cases where $x > 0$ (denoted $x^+$) and $x < 0$ (denoted $x^-$) which correspond to fixed values of $g$. This gives

\begin{equation}\label{eq:jointGplus}
    \begin{aligned}
        p(g=1, x^+) &=  \frac{1}{2\sqrt{2\pi}\sigma} \text{exp}\left( {-\frac{1}{2} \cdot \frac{(x^+ - \theta)^2}{\sigma^2} } \right) \\
        p(g=1, x^-) &=  \frac{1}{2\sqrt{2\pi}\sigma} \text{exp}\left( {-\frac{1}{2} \cdot \frac{(x^- + \theta)^2}{\sigma^2} } \right) \, ,
    \end{aligned}
\end{equation}

\begin{equation}\label{eq:jointGminus}
    \begin{aligned}
        p(g=-1, x^+) &=  \frac{1}{2\sqrt{2\pi}\sigma} \text{exp}\left( {-\frac{1}{2} \cdot \frac{(x^+ + \theta)^2}{\sigma^2} } \right) \\
        p(g=-1, x^-) &=  \frac{1}{2\sqrt{2\pi}\sigma} \text{exp}\left( {-\frac{1}{2} \cdot \frac{(x^- - \theta)^2}{\sigma^2} } \right) \, .
    \end{aligned}
\end{equation}

Since the Schmidt et al. Gaussian mixture is created symmetrically (the probability mass is evenly split between the two classes i.e., the mixture comprises one Gaussian offset by $\theta$ and mirrored at $x=0$) we can simplify to

\begin{equation}\label{eq:jointGplusMirror}
    \begin{aligned}
        p(g=1, x) &=  \frac{1}{\sqrt{2\pi}\sigma} \text{exp}\left( {-\frac{1}{2} \cdot \frac{(x - \theta)^2}{\sigma^2} } \right) ,
    \end{aligned}
\end{equation}

\begin{equation}\label{eq:jointGminusMirror}
    \begin{aligned}
        p(g=-1, x) &=  \frac{1}{\sqrt{2\pi}\sigma} \text{exp}\left( {-\frac{1}{2} \cdot \frac{(x + \theta)^2}{\sigma^2} } \right) ,
    \end{aligned}
\end{equation}

where $x > 0 ~~ \AND ~ x \sim \calN(\pm\theta, \sigma^2)$. In words, Equation~\ref{eq:jointGminusMirror} is the symmetrical tail of the Gaussian mixture while Equation~\ref{eq:jointGplusMirror} is the remainder of the mixture.

Similarly, a point-wise gradient is given as the Bernoulli outcome $g \in \{\pm 1\}$. The choice of $\epsilon$ directly influences the marginal probability over the manifold. The marginal probability over the manifold can be given generally as the Riemann approximations 

\begin{equation}\label{eq:marginalGplus}
    \begin{aligned}
        p(g=1)_\epsilon &= \frac{1}{\sqrt{2\pi}\sigma} \sum_{i=1}^{n}  \text{exp}\left( {-\frac{1}{2} \cdot \frac{(x_i^* - \theta)^2}{\sigma^2} } \right) \Delta_i,
    \end{aligned}
\end{equation}

and

\begin{equation}\label{eq:marginalGminus}
    \begin{aligned}
        p(g=-1)_\epsilon &= \frac{1}{\sqrt{2\pi}\sigma} \sum_{i=1}^{n}  \text{exp}\left( {-\frac{1}{2} \cdot \frac{(x_i^* + \theta)^2}{\sigma^2} } \right) \Delta_i ,
    \end{aligned}
\end{equation}

with $\Delta_i = x_{i} - x_{i-1}$ and for all positive $x_i^* \in [x_{i-1}, x_i]$ and $n$ is controlled by the hyper-parameter $\epsilon$. The $\epsilon$ is omitted when dealing with $n=1$, the non-robust case.

The marginal for the manifold under the gradient is given similarly as

\begin{equation}
    \begin{aligned}
        p(x) &=  \frac{1}{\sqrt{2\pi}\sigma} \text{exp}\left( {-\frac{1}{2} \cdot \frac{(x - \theta)^2}{\sigma^2} } \right) \\
        &+  \frac{1}{\sqrt{2\pi}\sigma} \text{exp}\left( {-\frac{1}{2} \cdot \frac{(x + \theta)^2}{\sigma^2} } \right) ,
    \end{aligned}
\end{equation}

where $x > 0$. Next denote the sub-manifold sampled from the positive ($y=1$) and negative ($y=-1$) classes as $\calM^+$ and $\calM^-$, respectively. 

Our definition for manifold-gradient mutual information is based on the standard definition of mutual information from information theory~\citep{cover_elements_2006},

\begin{equation}\label{eq:MInonrobust}
    I(\grads, \calM)_\epsilon = \int_\calM \int_\grads p_{\grads\calM}(g, x) \,\text{log}(\frac{p_{\grads\calM}(g, x)}{p_\grads(g)p_\calM(x)}) \, dg \, dx.
\end{equation}

where $\epsilon$ is treated as a hyper-parameter controlling the value of $n$ in $p_\calG(g)$. By substitution into Equation~\ref{eq:MInonrobust} we have

\begin{equation}
\begin{aligned}
    I(\grads, \calM)_\epsilon &= \int_\calM p(1, x) \,\text{log}(\frac{p(1, x)}{p_\calG(1)p_\calM(x)}) \, dx 
    + \int_\calM p(-1, x) \,\text{log}(\frac{p(-1, x)}{p_\calG(-1)p_\calM(x)}) \, dx
\end{aligned}
\end{equation}

This is split further similar to true positive, true negative, false positive, and false negative, as 

\begin{equation}
\begin{aligned}
    I(\grads, \calM)_\epsilon &= \int_{\calM^+} p(1, x^+) \,\text{log}(\frac{p(1, x^+)}{p_\calG(1)p_\calM(x^+)}) \, dx^+ \\
    &+ \int_{\calM^-} p(1, x^-) \,\text{log}(\frac{p(1, x^-)}{p_\calG(1)p_\calM(x^-)}) \, dx^- \\
    &+ \int_{\calM^+} p(-1, x^+) \,\text{log}(\frac{p(-1, x^+)}{p_\calG(-1)p_\calM(x^+)}) \, dx^+ \\
    &+ \int_{\calM^-} p(-1, x^-) \,\text{log}(\frac{p(-1, x^-)}{p_\calG(-1)p_\calM(x^-)}) \, dx^-,
\end{aligned}
\end{equation}

and simplified due to symmetry at 0 as
\begin{equation}\label{eq:MI:exact_app}
    \begin{aligned}
        I(\grads, \calM)_\epsilon &= 2\int_{\calM^+} p(1, x^+) \,\text{log}(\frac{p(1, x^+)}{p_\calG(1) p_\calM(x^+)}) \, dx^+ 
        + 2\int_{\calM^+} p(-1, x^+) \,\text{log}(\frac{p(-1, x^+)}{p_\calG(-1)p_\calM(x^+)}) \, dx^+.
    \end{aligned}
\end{equation}

Notably the cases for each possible scenario under detection theory are represented. 
\section{Implementation details}
\label{app:imp_details}

\subsection{Adversary Autoencoder}

We are primarily interested in the effect of reduced search resolution on attack behavior. Thus in this work, given a candidate direction $\vectheta'$ and magnitude (or radius) $r$, the adversarial sample in the AE case is the blending $(1-r)\vecx_0 + r\calD\left(\calE(\vecx_0) + \vectheta'\right)$.\footnote{We observed that it is detrimental to set $\vecx = \calD(\calE(\vecx_0) + r\vectheta')$ or $\vecx = \calD(r\vectheta')$ directly. Despite remaining on the data manifold by attacking it directly, the approximation of the data manifold is crude, which results in large distortion~\citep{Stutz_2019_CVPR}.}

For AE attack variants, we implement the same architecture described by~\citet{tu_autozoom_2019}. Specifically it leverages a fully convolutional network for the encoder and decoder. Every AE is trained using the held out test set, as we assume disjoint data between adversary and victim. 
\section{Supplemental Results}
\label{sec:supp_results}

\subsection{Attacks on ImageNet}
\label{app:imagenet}
We provide supplemental results on the ImageNet~\citep{ILSVRC15} classification dataset. Ten random classes are chosen with ten random samples taken from each (100 total samples). The natural architecture is the pre-trained Resnet50 network taken from the PyTorch Torchvision library.\footnote{\url{https://pytorch.org/docs/stable/torchvision/models.html}} For the robust case, we compare against the SotA at time of writing, randomized smoothing proposed by~\citet{cohen_certified_2019}. We use the pre-trained Resnet50 weights and implementation provided by Cohen et al., corresponding to smoothing parameter $\sigma=0.5$ and $\epsilon \simeq 1.0$. When performing attacks on ImageNet, we use the attack's respective $L_{2}$-norm version, since randomized smoothing was certified under $L_{2}$-norm setting. ImageNet samples are downsized to 128x128 before passing to the AE, and the output of the AE is scaled back to 224x224, as described by~\citet{tu_autozoom_2019}.

\begin{figure*}[t!]
    \centering
\subfloat[]{
    \includegraphics[scale=1.0,trim={0 0 0 0},clip]{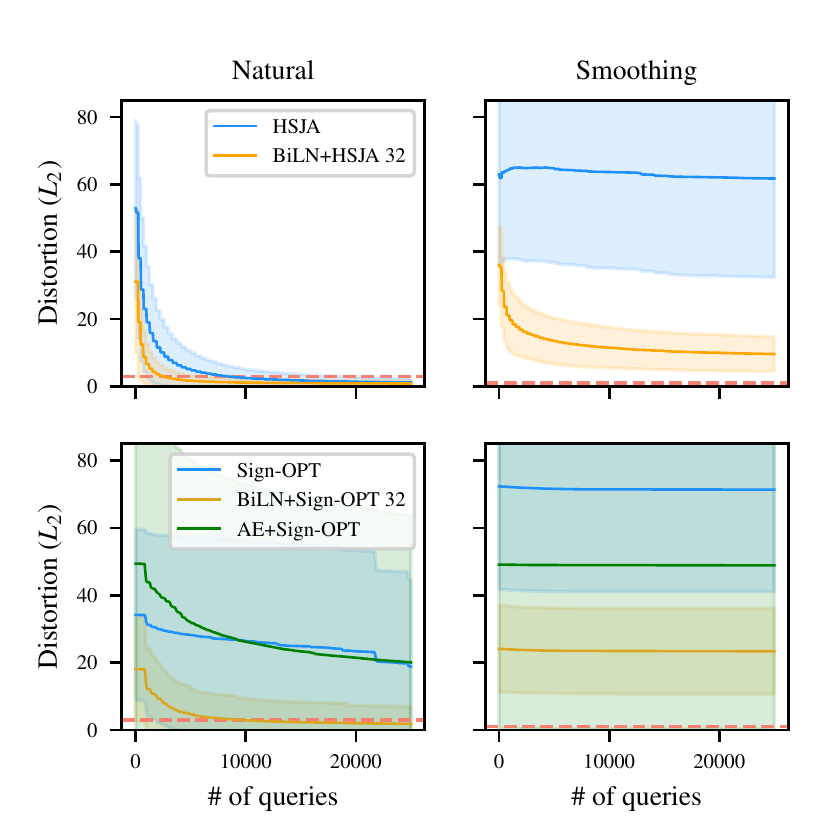}} 
% \hspace{0.2em}
\subfloat[]{
    \includegraphics[scale=1.0,trim={0 0 0 0},clip]{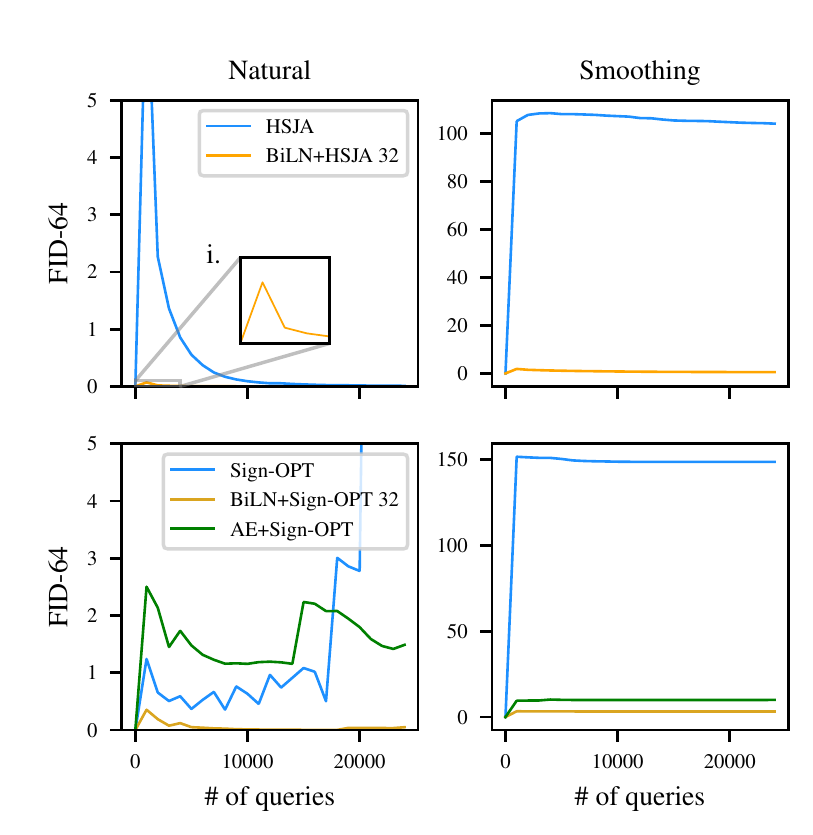}}\\
\vspace{0em}

% \vspace{-0.7em}
    \caption{Results across attacks for ImageNet dataset, corresponding to a) distortion against query usage (dotted red line denotes the value of $\epsilon$, shaded areas mark standard deviation), and b) FID-64 trajectory against the same query usage.}
    \label{fig:imagenet}
\end{figure*}

\paragraph{ImageNet case study ($L_2$).} We attack ImageNet in the $L_2$-norm setting to compare against the certified smoothing technique proposed by~\citet{cohen_certified_2019}. The label output comes from a smooth classifier, approximated by many rounds of Monte Carlo search, which uses the regular model regularized by Gaussian noise. Notably the smoothing occurs at inference, so there is no increase in the number of training points. The distortion results of these attacks are shown in Figure~\ref{fig:imagenet}a. Dimension reduction has a larger impact when coupled with the large ImageNet resolution. Particularly the BiLN+HSJA (orange line, top row) and BiLN+Sign-OPT (orange line, bottom row) attacks profit the most. At 8k queries, success rate increases 1.4x and 2.1x for HSJA and Sign-OPT, respectively. This is due to 1) the AE only providing a crude approximation of the ImageNet manifold, by only having access to the test set, and 2) BiLN allowing to search closer to the original sample, since it is a deterministic function independent of the adversary's knowledge. 

% As before with CIFAR-10, the RayS dimension reduction is saturated on ImageNet (Insets~\ref{fig:imagenet}a.i and \ref{fig:imagenet}a.ii). Since RayS does not explicitly rely on gradient estimation, it benefits the least from dimension-reduction techniques. RayS saw no such improvement, as it does not perform an explicit gradient estimate (Inset~\ref{fig:imagenet}b.i).

The FID scores in Figure~\ref{fig:imagenet}b paint a more comprehensive picture. BiLN variants (orange lines) produce adversarial examples closer to the manifold than either regular (blue) or AE (greeN) variants, highlighted with HSJA+BiLN in Inset~\ref{fig:imagenet}b.i.  We interpret this as follows: BiLN variants on HSJA and Sign-OPT leverage reduced dimensionality to increase the manifold-gradient mutual information, and 1) produce a smoother noise distribution, resulting in more spatially correlated distortion, which as a result 2) produces adversarial examples closer to the manifold. Another key observation is the fluctuation of LID score towards the end of Sign-OPT and AE+Sign-OPT, which are not present for HSJA (first column of Figure~\ref{fig:imagenet}b). Notably there is no direct signal of manifold distance in the experiments, so the adversary relies on implicit manifold distance feedback from the model, which can be inaccurate.

\subsection{Attacking without gradient estimate}
\label{app:rays}
We perform additional experiments with an attack that does not perform an explicit gradient estimate. 

\paragraph{RayS.} \citet{chen_rays_2020} propose an alternative hard-label attack method which is to search for the minimum decision boundary radius $r$ from a sample $\vecx_0$, along a ray direction $\vectheta$. Instead of searching over $\reals^d$ to minimize $g(\vectheta)$, Chen et al. propose to perform ray search over directions $\vectheta \in \{-1, 1\}^d$, resulting in $2^d$ maximum possible directions. This reduction of the search resolution enables SotA query efficiency in the $L_{\infty}$ setting with proof of convergence. The search resolution is further reduced by the hierarchical variant of RayS, which performs on-the-fly upscaling of image super-pixels.

The intuition behind RayS attack is to perform a discrete search in at most $2^d$ directions. Chen et al. also perform a hierarchical search over progressively larger super-pixels of the image. This has the effect of already upscaling \textit{on-the-fly}~\citep{chen_rays_2020}. RayS has the unique behavior of performing a discrete search for the decision boundary, rather than an explicit gradient estimate. To achieve an appropriate reduced-dimension version of RayS, we modify the calculation of $s$ in Algorithm 3 of \citet{chen_rays_2020}, which either speeds up upscaling by a factor $a$ (i.e., $s = s + a$), or extends the search through a specific block index by a factor $b$ (increase block level at $k = 2^sb$ instead of $k = 2^s$). 

\begin{figure*}[t!]
    \centering
\subfloat[]{
    \includegraphics[scale=1.0,trim={0 0 0 0},clip]{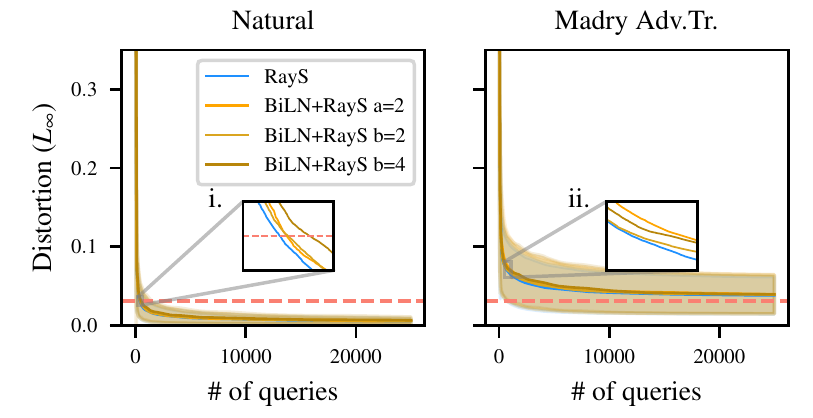}} 
% \hspace{0.2em}
\subfloat[]{
    \includegraphics[scale=1.0,trim={0 0 0 0},clip]{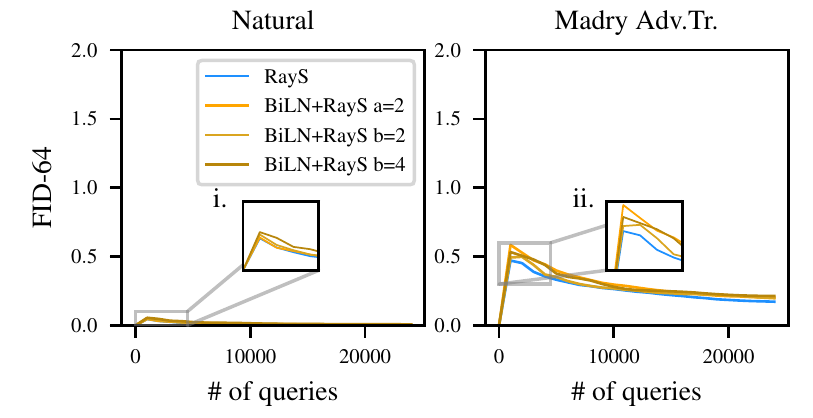}}\\
\vspace{0em}

% \vspace{-0.7em}
    \caption{Results for RayS on the CIFAR-10 dataset, corresponding to a) distortion against query usage (dotted red line denotes the value of $\epsilon$, shaded areas mark standard deviation), and b) FID-64 trajectory against the same query usage. }
    \label{fig:cifar10_rays}
\end{figure*}

\paragraph{Results.} The result of attacking CIFAR-10 with RayS is shown in Figure~\ref{fig:cifar10_rays}. The BiLN variants of RayS each have minimal effect on overall query efficiency (Insets~\ref{fig:cifar10_rays}a.i and \ref{fig:cifar10_rays}a.ii). This is a result of RayS not relying on explicit gradient estimation. When comparing the FID-64 score, the dimension-reduced variants of RayS do not have a large variation between them (Inset~\ref{fig:cifar10_rays}b.i), a side-effect of the adaptive super-pixel search, which can automatically scale the super-pixel size as the search progresses. 

\subsection{Success Rate Plots}
\label{app:sr_plots}

\begin{figure}[h!]
    \centering
    \includegraphics[scale=0.97,trim={0 0 0 0},clip]{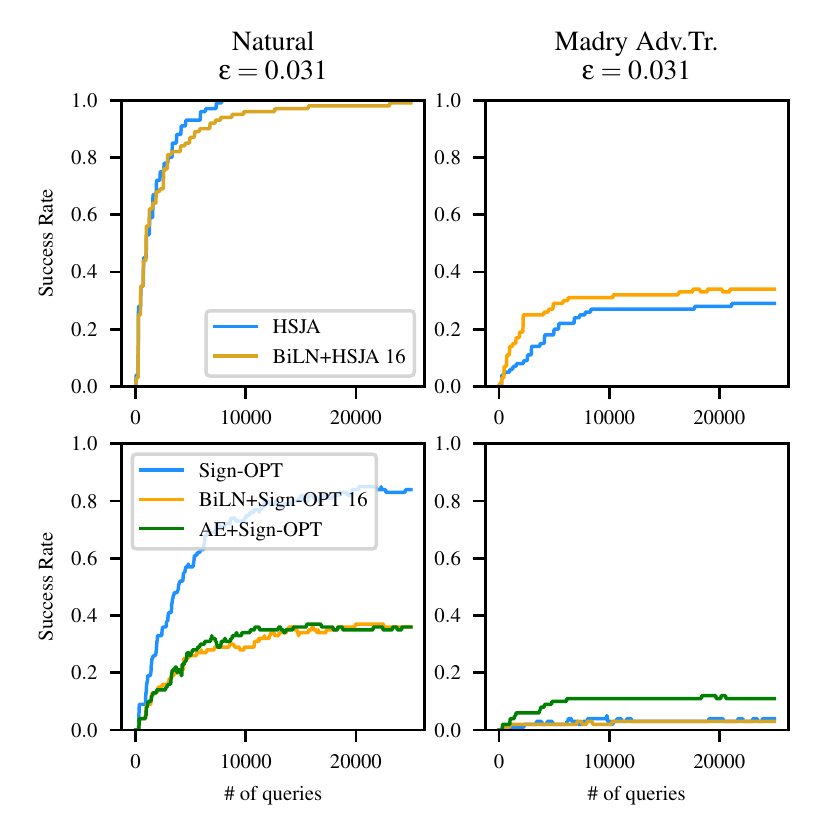}
% \vspace{-0.7em}
    \caption{Query vs. success rate plots corresponding to each attack variant in the main text on CIFAR-10 . }
    \label{fig:sr_figs}
\end{figure}

In Figure~\ref{fig:sr_figs} we provide query vs. success rate to accompany the results in the main text.

\subsection{Tabular results - CIFAR-10}
\label{app:table_results_cifar10}

In Table~\ref{app:cifar_table} we provide tabular comparison at certain query intervals between regular and robust models for CIFAR-10.

\begin{table*}[]
\centering
% \resizebox{\textwidth}{!}{%
\begin{tabular}{@{}lllllllll@{}}
\toprule
	&	\multicolumn{4}{l}{CIFAR-10 ($\epsilon=0.031$) } &	\multicolumn{4}{l}{Madry Adv.Tr. ($\epsilon=0.031$) }	\\ 
	 &	\# Queries &	Avg. $L_{\infty}$ &	SR &	FID-64 &	\# Queries &	Avg. $L_{\infty}$ &	SR &	FID-64	\\ \midrule
\multirow{3}{*}{RayS}	&	4,000	&	0.01	&	98.0	&	0.02	&	4,000	&	0.05	&	33.0	&	0.36	\\ 
	&	8,000	&	0.01	&	100.0	&	0.01	&	8,000	&	0.04	&	36.0	&	0.28	\\ 
	&	14,000	&	0.01	&	100.0	&	0.01	&	14,000	&	0.04	&	39.0	&	0.23	\\ \bottomrule 
\multirow{3}{*}{BiLN+RayS a=2}	&	4,000	&	0.01	&	99.0	&	0.02	&	4,000	&	0.05	&	32.0	&	0.44	\\ 
	&	8,000	&	0.01	&	100.0	&	0.01	&	8,000	&	0.05	&	35.0	&	0.33	\\ 
	&	14,000	&	0.01	&	100.0	&	0.01	&	14,000	&	0.04	&	36.0	&	0.25	\\ \bottomrule 
\multirow{3}{*}{BiLN+RayS b=2}	&	4,000	&	0.01	&	100.0	&	0.02	&	4,000	&	0.05	&	33.0	&	0.37	\\ 
	&	8,000	&	0.01	&	100.0	&	0.01	&	8,000	&	0.04	&	37.0	&	0.29	\\ 
	&	14,000	&	0.01	&	100.0	&	0.01	&	14,000	&	0.04	&	39.0	&	0.24	\\ \bottomrule 
\multirow{3}{*}{BiLN+RayS b=4}	&	4,000	&	0.01	&	93.0	&	0.03	&	4,000	&	0.05	&	32.0	&	0.44	\\ 
	&	8,000	&	0.01	&	100.0	&	0.02	&	8,000	&	0.05	&	35.0	&	0.33	\\ 
	&	14,000	&	0.01	&	100.0	&	0.01	&	14,000	&	0.04	&	38.0	&	0.25	\\ \bottomrule 
\multirow{3}{*}{HSJA}	&	4,000	&	0.01	&	88.0	&	0.06	&	4,000	&	0.09	&	15.0	&	3.57	\\ 
	&	8,000	&	0.01	&	100.0	&	0.02	&	8,000	&	0.08	&	26.0	&	2.41	\\ 
	&	14,000	&	0.01	&	100.0	&	0.01	&	14,000	&	0.06	&	27.0	&	1.69	\\ \bottomrule 
\multirow{3}{*}{BiLN+HSJA 16}	&	4,000	&	0.02	&	82.0	&	0.03	&	4,000	&	0.07	&	25.0	&	0.68	\\ 
	&	8,000	&	0.01	&	94.0	&	0.01	&	8,000	&	0.06	&	31.0	&	0.47	\\ 
	&	14,000	&	0.01	&	97.0	&	0.01	&	14,000	&	0.05	&	32.0	&	0.37	\\ \bottomrule 
\multirow{3}{*}{Sign-OPT}	&	4,000	&	0.06	&	52.0	&	0.70	&	4,000	&	0.33	&	2.0	&	2.28	\\ 
	&	8,000	&	0.04	&	71.0	&	0.89	&	8,000	&	0.30	&	3.0	&	1.04	\\ 
	&	14,000	&	0.02	&	79.0	&	0.00	&	14,000	&	0.28	&	3.0	&	0.59	\\ \bottomrule 
\multirow{3}{*}{BiLN+Sign-OPT 16}	&	4,000	&	0.09	&	21.0	&	0.01	&	4,000	&	0.37	&	2.0	&	0.08	\\ 
	&	8,000	&	0.06	&	29.0	&	0.00	&	8,000	&	0.35	&	3.0	&	0.04	\\ 
	&	14,000	&	0.06	&	36.0	&	0.00	&	14,000	&	0.33	&	3.0	&	0.04	\\ \bottomrule 
\multirow{3}{*}{AE+Sign-OPT}	&	4,000	&	0.09	&	20.0	&	0.01	&	4,000	&	0.20	&	8.0	&	0.01	\\ 
	&	8,000	&	0.07	&	31.0	&	0.00	&	8,000	&	0.17	&	11.0	&	0.21	\\ 
	&	14,000	&	0.06	&	35.0	&	0.00	&	14,000	&	0.15	&	11.0	&	0.21	\\ \bottomrule 
\end{tabular}
% }
\caption{Comparison at certain query intervals between regular and robust models on CIFAR-10.}
\label{app:cifar_table}
\end{table*}

\subsection{Visual results - CIFAR-10}
\label{app:visual_cifar10}

\begin{figure*}[h!]
    \centering
    \subfloat[Sign-OPT on CIFAR10]{
        \includegraphics[scale=0.97]{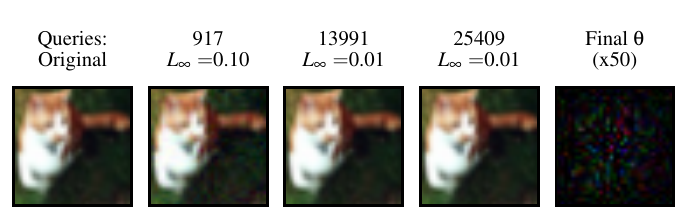}}
    \hspace{0.2em}
    \subfloat[Sign-OPT on CIFAR10 (Madry Adv. Tr.)]{
        \includegraphics[scale=0.97]{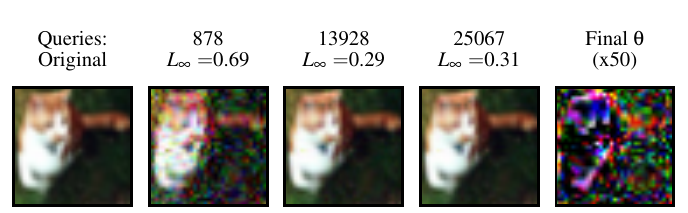}}\\
    \vspace{0em}
    
    \subfloat[BiLN Sign-OPT on CIFAR10]{
        \includegraphics[scale=0.97]{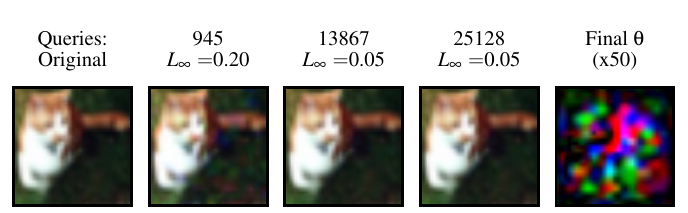}}
    \hspace{0.2em}
    \subfloat[BiLN Sign-OPT on CIFAR10 (Madry Adv. Tr.)]{
        \includegraphics[scale=0.97]{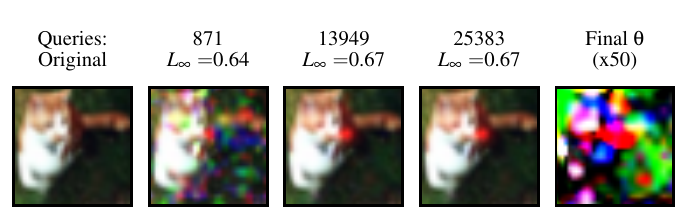}}\\
    \vspace{0em}
    
    \subfloat[AE Sign-OPT on CIFAR10]{
        \includegraphics[scale=0.97]{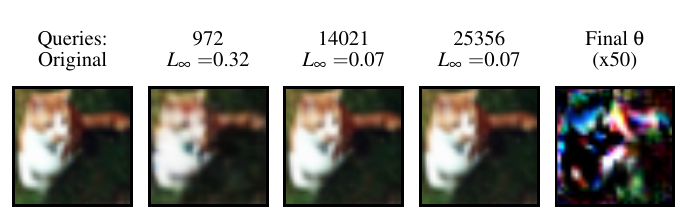}}
    \hspace{0.2em}
    \subfloat[AE Sign-OPT on CIFAR10 (Madry Adv. Tr.)]{
        \includegraphics[scale=0.97]{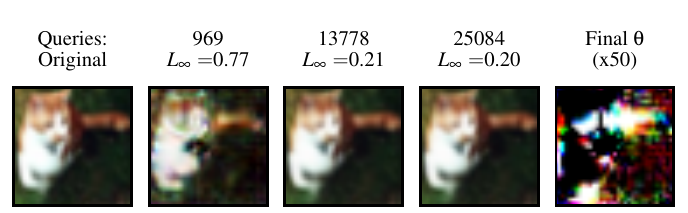}}\\
    \vspace{0em}
    
    \subfloat[HSJA on CIFAR10]{
        \includegraphics[scale=0.97]{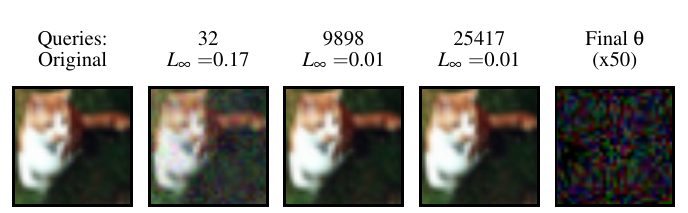}}
    \hspace{0.2em}
    \subfloat[HSJA on CIFAR10 (Madry Adv. Tr.)]{
        \includegraphics[scale=0.97]{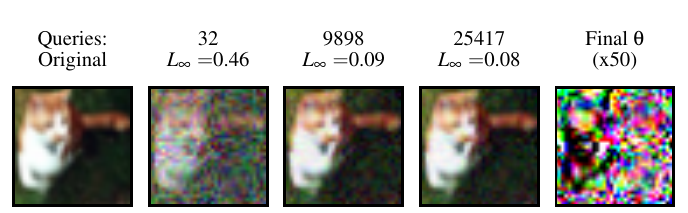}}\\
    \vspace{0em}
    
    \subfloat[BiLN HSJA on CIFAR10]{
        \includegraphics[scale=0.97]{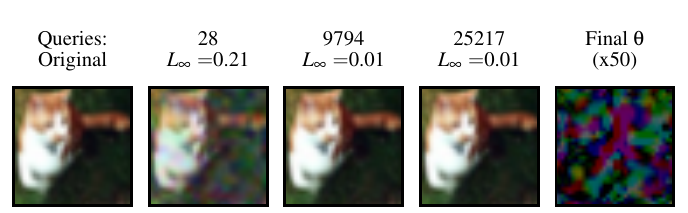}}
    \hspace{0.2em}
    \subfloat[BiLN HSJA on CIFAR10 (Madry Adv. Tr.)]{
        \includegraphics[scale=0.97]{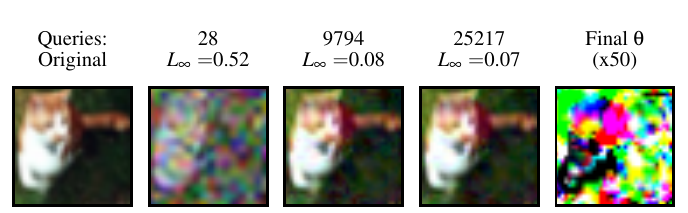}}\\
    \vspace{0em}
    
    \subfloat[RayS on CIFAR10]{
        \includegraphics[scale=0.97]{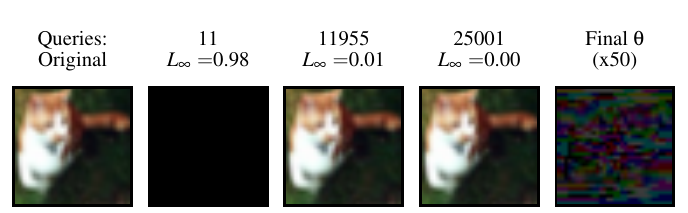}}
    \hspace{0.2em}
    \subfloat[RayS on CIFAR10 (Madry Adv. Tr.)]{
        \includegraphics[scale=0.97]{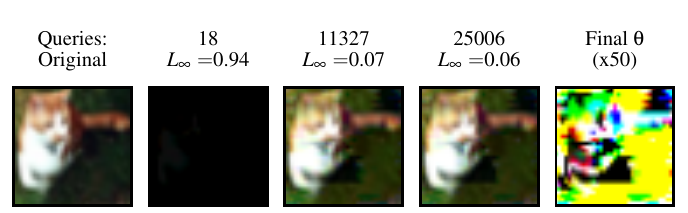}}

% \vspace{-0.7em}
    \caption{Visual selection of attack trajectories on CIFAR-10.}
\label{fig:visual_cifar10}
\end{figure*}

We provide visual qualitative results for each attack on CIFAR-10 in Figure~\ref{fig:visual_cifar10}.

\end{document}